\documentclass[3p,times,twocolumn]{elsarticle}
\usepackage{amssymb}
\usepackage{hyperref}
\hypersetup{
    colorlinks=true,
    linkcolor=black,
    citecolor=black,
    a4paper=true,
    plainpages=false}
\usepackage{xspace}
\usepackage{epsfig}
\usepackage{graphicx}
\usepackage{amsmath}
\usepackage[caption=false]{subfig}
\usepackage{color}
\usepackage[utf8]{inputenc}
\usepackage{enumitem}

\graphicspath{{images/}}

\makeatletter
\DeclareRobustCommand\onedot{\futurelet\@let@token\@onedot}
\def\@onedot{\ifx\@let@token.\else.\null\fi\xspace}
\def\eg{\emph{e.g}\onedot} 
\def\ie{\emph{i.e}\onedot} 
\def\cf{\emph{cf}\onedot} 
 \def\vs{\emph{vs}\onedot}
\def\wrt{w.r.t\onedot} 
\def\etal{\emph{et al}\onedot}
\def\vs{vs\onedot}
\makeatother

\newcommand{\PAR}[1]{\vskip4pt \noindent {\bf #1~}}

\journal{arXiv}

\begin{document}

\begin{frontmatter}

\title{Visual Landmark Recognition from Internet Photo Collections:\\A Large-Scale Evaluation}

\author{Tobias Weyand\corref{cor1}}
\ead{weyand@vision.rwth-aachen.de}

\author{Bastian Leibe}
\ead{leibe@vision.rwth-aachen.de}

\address{Computer Vision Group\\RWTH Aachen University\\Germany}

\begin{abstract}
  The task of a visual landmark recognition system is to identify
  photographed buildings or objects in query photos and to provide the
  user with relevant information on them. With their increasing
  coverage of the world's landmark buildings and objects, Internet
  photo collections are now being used as a source for building such
  systems in a fully automatic fashion. This process typically
  consists of three steps: clustering large amounts of images by the
  objects they depict; determining object names from user-provided
  tags; and building a robust, compact, and efficient recognition
  index. To this date, however, there is little empirical information
  on how well current approaches for those steps perform in a
  large-scale open-set mining and recognition task. Furthermore, there
  is little empirical information on how recognition performance
  varies for different types of landmark objects and where there is
  still potential for improvement. With this paper, we intend to fill
  these gaps. Using a dataset of 500k images from Paris, we analyze
  each component of the landmark recognition pipeline in order to
  answer the following questions: How many and what kinds of objects
  can be discovered automatically? How can we best use the resulting
  image clusters to recognize the object in a query? How can the
  object be efficiently represented in memory for recognition? How
  reliably can semantic information be extracted? And finally: What
  are the limiting factors in the resulting pipeline from query to
  semantics?  We evaluate how different choices of methods and
  parameters for the individual pipeline steps affect overall system
  performance and examine their effects for different query categories
  such as buildings, paintings or sculptures.
\end{abstract}


\end{frontmatter}

\section{Introduction}
\label{sec:intro}

Recognizing the object in a photo is one of the fundamental problems
of computer vision. One generally distinguishes between object
categorization and specific object recognition. Object categorization
means recognizing the \emph{class} that an object belongs to, \eg
painting or building, while specific object recognition means
recognizing a specific object \emph{instance}, such as the Mona Lisa
or the Eiffel Tower. In this paper, we consider the latter task, \ie,
specific object recognition. In particular, we are interested in two
applications, namely photo auto-annotation and mobile visual search. A
photo auto-annotation system recognizes objects in a user's photo
albums and labels them automatically, saving the user the effort of
manually labeling them. A mobile visual search system provides a user
with information on an object that they took a picture of with their
smartphone. Because a large part of the photos in these applications
are typically tourist photos, many of the objects that such systems
need to recognize are landmarks. Therefore, the problem is typically
referred to as \emph{landmark recognition}. However, many other types
of objects, such as paintings, sculptures or murals, can also be
recognized by such systems.

The first step of building a landmark recognition system is to compile
a database consisting of one or more photos of each object that shall
be recognized. However, since the number of objects that can possibly
appear in a user's photos is virtually infinite, it is impossible to
construct and maintain such a database by hand. An elegant solution is
to build the database from the data it is meant to be applied to,
namely public photos from Internet photo collections such as Flickr,
Picasa or Panoramio.  This approach has several attractive properties:
(i) Objects are discovered in an unsupervised, fully automatic way,
making it unnecessary to manually create a list of objects and
collecting photos for each of them. (ii) The resulting set of objects
is likely to be much better adapted to the queries a photo
auto-annotation or visual search system might receive than a
hand-collected set of objects. (iii) The level of detail of object
representation is automatically adapted to the demand.  The most
popular objects will be represented by the most photos in the
database, increasing their chance of successful recognition, while
only little memory is used on less popular objects. This approach has
gained popularity in the research community
\cite{Avrithis10MM,Gammeter09ICCV,Kalantidis11MMTA,Quack08CIVR,Weyand11ICCV}
and is also being used in applications such as Google Goggles
\cite{Zheng09CVPR}.

Constructing such landmark recognition systems on a large scale
involves three main research problems: (i) Finding interesting
structures in internet image collections, (ii) automatically
connecting them to associated semantic content by examining
user-provided titles and tags, and (iii) compactly representing them
for efficient retrieval. While each of those problems has already been
studied in isolation, so far there has not been a systematic
evaluation of all three aspects in the context of a fully automatic
pipeline.
Image retrieval approaches like \cite{Chum07ICCV,Philbin07CVPR} find
matching images for a query, but have no notion of the semantics of
the depicted content. Tag mining approaches
\cite{Kalantidis11MMTA,Crandall09WWW,Quack08CIVR,Simon07ICCV} try to
find a description of an image cluster, but so far the large effort to
evaluate tag quality has prevented quantitative evaluations in a
large-scale setting. Landmark object discovery approaches
\cite{Avrithis10MM,Frahm10ECCV,Quack08CIVR,Weyand11ICCV,Zheng09CVPR}
aim at finding interesting buildings and other objects, but no
systematic evaluation has been performed that analyzes what types of
objects can be discovered and how recognition performance varies with
these types. Furthermore, it is still largely unclear what is the best
strategy to determine the identity of the recognized object based on
the set of retrieved database images (which becomes a non-trivial
problem whenever image clusters overlap and images may contain
multiple landmark objects).

In this paper, we evaluate the whole process of constructing landmark
recognition engines from Internet photo collections. To do this in a
realistic large-scale setting we require a dataset containing
thousands of objects.  Moreover, in order to create a realistic
application scenario, the target database objects should not be
specified by hand (as in many other datasets), but should be mined
automatically. Last but not least, the dataset should not be limited
to buildings, but also contain smaller objects, such as paintings or
statues.

Our evaluation is based on the \textsc{Paris 500k} dataset
\cite{Weyand10RMLE} containing 500k photos from the inner city of
Paris, which was mined from Flickr and Panoramio using a geographic
bounding region rather than keyword queries to obtain a distribution
unbiased towards specific landmarks.  Thus, in contrast to other
common datasets, there is no bias on tag annotations or content. In
order to evaluate landmark recognition in a realistic setting, we
additionally collected a query set of almost 3,000 Flickr images from
Paris that is disjoint from the original dataset. Our evaluation thus
mimics the task of photo auto-annotation where a photo uploaded to a
photo sharing website is automatically annotated with the object it
depicts.

To evaluate the performance of landmark recognition, we use a recent
landmark discovery algorithm \cite{Weyand11ICCV} to discover landmarks
in the dataset. We created an exhaustive ground truth for the
relevance of each of the discovered landmarks with respect to each of
the 3,000 queries, which involved significant manual effort. This is
the first ground truth for evaluating landmark recognition on an
unbiased and realistic dataset. To enable the comparison of other
approaches with the ones evaluated in this paper, the ground truth is
publicly available.

To give a detailed performance analysis for different types of
objects, we introduce a taxonomy for the objects landmark recognition
systems are able to recognize.  Throughout our evaluation, we report
both summary performances over the entire database and detailed
findings for different object categories that show how their
recognition is affected by the different stages of the system.  As our
results show, the observed effects vary considerably between query
categories, justifying this approach. We give detailed results for
each category, and use the four use cases of \emph{Landmark
  Buildings}, \emph{Paintings}, \emph{Building Details} and
\emph{Windows} as representatives for different challenges. The
taxonomy is available along with the ground truth.

Note that our goal is not primarily to propose novel methods (although
some of the methods evaluated in Sec.~\ref{sec:recognition} and
Sec.~\ref{sec:retrieval} are indeed novel), but to provide answers to
the following questions:
\begin{itemize}[noitemsep]
\item How many and what kinds of objects
are present in Internet photo collections and what is the difficulty
of discovering objects of different landmark types
(Sec.~\ref{sec:landmarks})?
\item   How to decide which landmark was
recognized given a list of retrieved images
(Sec.~\ref{sec:recognition})?
\item How to efficiently represent the
discovered objects in memory for recognition
(Sec.~\ref{sec:retrieval})?
\item Are the user-provided tags reliable
enough for determining accurate object names
(Sec.~\ref{sec:semantics})?
\item Given the entire retrieval, recognition and semantic labeling
  pipeline, what are the factors effectively limiting the recognition
  of different object categories (Sec.~\ref{sec:end_to_end})?
\end{itemize}

Our analysis provides several interesting insights, for example:

\begin{itemize}[noitemsep]
\item Semantic annotation is the main bottleneck for system
  performance. In many cases, the correct object is visually
  recognized, but the name of the object cannot be determined due to
  the sparsity and amount of noise of user-provided image titles and
  tags.
\item Different bottlenecks exist for different object categories. For
  example, \emph{Murals} are easy to recognize using the standard
  visual words pipeline, but reliable semantic information is often
  missing for them. For other objects like museum exhibits, the
  opposite is the case: While semantic information is readily
  available, they are hard to recognize visually due to their spatial
  structure and scarce visual examples.
\item When the desired application is building recognition, a
  seeding-based clustering method can bring significant computational
  savings, since buildings are already discovered when using few
  seeds, while smaller objects require orders of magnitude more seeds.
\item Different techniques for compactly representing object clusters
  are optimal for different object types.
\end{itemize}

As a result of this evaluation, we can identify several interesting
directions, where progress can still be made.

\section{Engine Architecture}
\label{sec:framework}
\begin{figure*}[t]
  \centering  \includegraphics[width=\linewidth]{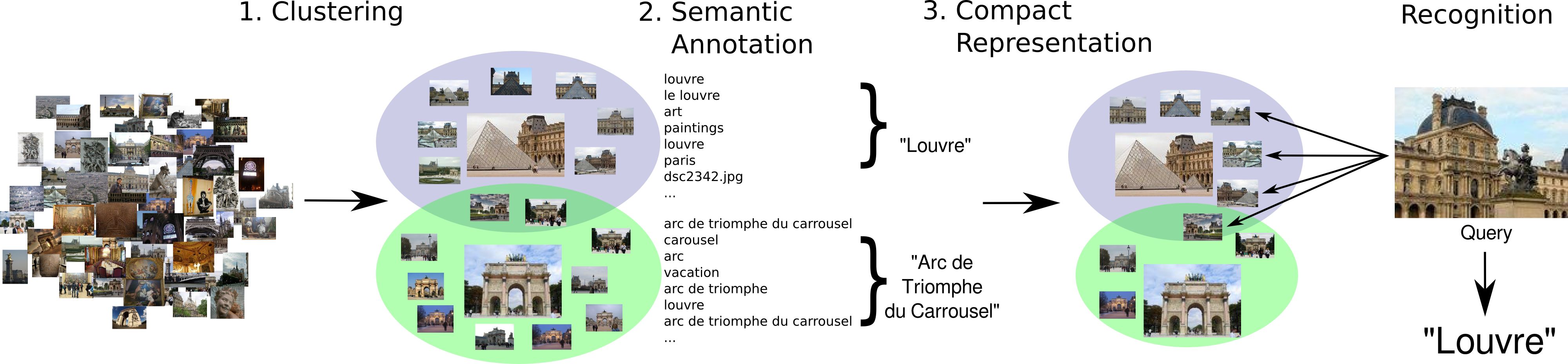}
  \caption{The architecture of a landmark recognition
    system. 1.~\emph{Objects} are discovered by visually clustering
    touristic photos (Sec.~\ref{sec:landmarks}). We call the photos in
    an object cluster its \emph{representatives}. 2.~\emph{Semantic}
    annotations are mined from user-provided tags
    (Sec.~\ref{sec:semantics}). 3.~The search index is made more
    compact by eliminating redundancy
    (Sec.~\ref{sec:retrieval}). 4.~The object in a query photo is
    recognized by retrieving similar photos and exploiting the
    knowledge of their cluster memberships
    (Sec.~\ref{sec:recognition}).}
  \label{fig:retrieval_framework}
\end{figure*}
The architecture of a typical landmark recognition engine such as
\cite{Avrithis10MM,Gammeter10ECCV,Quack08CIVR,Zheng09CVPR} is shown in
Fig.~\ref{fig:retrieval_framework}. Large amounts of tourist photos
are clustered, resulting in a set of \emph{objects}. By \emph{object},
we denote a cluster of images that show the same entity. We will refer
to the images in each object cluster as its
\emph{representatives}. Since the clusters may overlap, a
representative can belong to multiple objects. Each object is then
associated with \emph{semantics} (typically its name), \eg, by mining
frequently used image tags. The set of representatives for each
cluster is often decimated by eliminating redundant images in order to
save memory and computation time. To recognize the object in a query
image, a visual search index
\cite{Nister2006CVPR,Philbin07CVPR,Sivic03ICCV} containing all
\emph{representatives} is queried, producing a ranked list of
matches. Based on this list, \emph{objects} are ranked \wrt their
relevance to the query and the corresponding \emph{semantics} are
returned.

In this paper, we evaluate different choices for the components of
this framework and demonstrate how they affect the system's overall
performance. Sec.~\ref{sec:landmarks} considers the stage of
determining a set of objects by clustering images from internet photo
collections and shows how many objects from which categories can be
discovered. Given a ranking of the representatives for a query,
Sec.~\ref{sec:recognition} analyzes different schemes for determining
the object shown in the query image. In Sec.~\ref{sec:retrieval} we
consider different ways of speeding up search and reducing memory
requirements by removing redundant representatives. Finally, in
Sec.~\ref{sec:semantics} we analyze the stage of semantic annotation
based on frequent tags and perform an end-to-end analysis of the
performance of the whole pipeline from query to semantics.

\section{Related Work}
\label{sec:related_work}
We now give an overview how the individual parts of the pipeline
introduced above have been approached in previous work.

\subsection{Datasets}
We created our own query set and ground truth for this paper, because
available benchmarks do not support such an evaluation. Most datasets
only cover very few, mostly building-scale, landmarks (\eg,
\textsc{European Cities 1M} \cite{Avrithis10MM}, \textsc{Statue of
  Liberty}, \textsc{Notre Dame} and \textsc{San Marco}
\cite{Li08ECCV}, \textsc{Oxford Buildings} \cite{Philbin07CVPR},
\textsc{Paris Buildings} \cite{Philbin08CVPR}). Another problem is
that their ground truths are designed for other tasks. Image retrieval
datasets (\eg \textsc{Oxford Buildings} \cite{Philbin07CVPR},
\textsc{Paris Buildings} \cite{Philbin08CVPR}, \textsc{INRIA Holidays}
\cite{Jegou08ECCV}) are not suitable for our evaluation, because we
want to evaluate object recognition, \ie recognizing the object(s) in
a query image, and not image retrieval, \ie retrieving images similar
to a query from a database. Image-based localization datasets
(\textsc{Aachen} \cite{Sattler12BMVC}, \textsc{Vienna}
\cite{Irschara09CVPR}, \textsc{Dubrovnik} and \textsc{Rome}
\cite{Li10ECCV}) evaluate how accurately the camera pose of the query
image can be estimated. While this is more related to our problem, our
goal differs from pose estimation, because camera pose does not
necessarily determine what object the camera is really seeing (See
Sec.~\ref{sec:relwork:lmrec} for more details.)
The \textsc{San Francisco} \cite{Baatz10ECCV} and \textsc{Landmarks
  1K} \cite{Li12ECCV} datasets are closest to our requirements, but
both of them focus on large, building-level landmarks while we are
explicitly interested in also evaluating the recognition of smaller,
non-building objects.

\subsection{Landmark Discovery}
\label{sec:relwork:lmdiscovery}
Landmark Recognition Engines are typically based on a
visual search index from objects discovered in Internet photo
collections \cite{Avrithis10MM,Gammeter10ECCV,Quack08CIVR,Zheng09CVPR}. The underlying \emph{landmark discovery} approaches
perform visual
\cite{Chum10PAMI,Li08ECCV,Philbin08ICCVGIP,Quack08CIVR,Weyand11ICCV,Johns11ICCV}
or geographical \cite{Crandall09WWW,Li09ICCV} clustering, or a combination of
the two \cite{Avrithis10MM,Zheng09CVPR,Ji12IJCV}. Zheng \etal
\cite{Zheng09CVPR} show that online tourist guides can be a valuable
additional data source, and Gammeter \etal \cite{Gammeter10ECCV} use
descriptions determined from user-provided tags to search for
additional images on the web. In this work however, we focus on
methods based solely on the images from Internet photo collections and their metadata.

Chum \etal \cite{Chum10PAMI} use \emph{Min-Hash} to find \emph{seed}
images and grow landmark clusters by query expansion
\cite{Chum07ICCV}.  Philbin \etal \cite{Philbin08ICCVGIP} over-segment
the matching graph using spectral clustering and merge clusters of the
same object based on image overlap. Gammeter \etal
\cite{Gammeter09ICCV} and Quack \etal \cite{Quack08CIVR} perform
hierarchical agglomerative clustering in a local matching
graph. Avrithis \etal \cite{Avrithis10MM} use Kernel Vector
Quantization to create a clustering with an upper bound on
intra-cluster dissimilarity. \emph{Iconoid Shift} by Weyand \etal
\cite{Weyand11ICCV} finds popular objects at different scales using
mode search based on a homography overlap distance. We choose Iconoid
Shift as our analysis tool, because it produces an overlapping
clustering and discovers landmarks at varying levels of granularity,
thus also discovering, \eg, building details.

\subsection{Landmark Recognition}
\label{sec:relwork:lmrec}
Based on the clusters resulting from landmark discovery, it is now possible to recognize the landmark in a query image. There are three predominant approaches for this in the literature: Image Retrieval, Classification and Pose Estimation.

\PAR{Image Retrieval.}%
Most \emph{Image Retrieval} based approaches use efficient \emph{specific object retrieval} methods \cite{Nister2006CVPR,Philbin07CVPR,Sivic03ICCV} that allow searching for images matching a query in a database consisting of potentially millions of images. 
Several approaches \cite{Quack08CIVR,Zheng09CVPR,Gammeter09ICCV,Philbin07CVPR,Sivic03ICCV}
implement a \emph{best match} strategy (Sec.~\ref{subsec:retrieval_methods}) where the query is matched against the database of representatives and the object cluster corresponding to the best match is returned. While Quack \etal \cite{Quack08CIVR} and Zheng \etal \cite{Zheng09CVPR} perform a precise but computationally expensive direct feature matching, Gammeter \etal \cite{Gammeter09ICCV} retrieve images using inverted indexing and bags-of-visual-words (BoVWs) \cite{Philbin07CVPR,Sivic03ICCV}.
Li \etal \cite{Li08ECCV} only want to decide \emph{whether} the query image contains a specific landmark. Given a dataset of photos of one landmark, they perform image retrieval based on both Gist features and BoVWs and apply a threshold to the retrieval score to decide if the query contains the object.
Both Avrithis \etal \cite{Avrithis10MM} and Johns \etal \cite{Johns11ICCV} compress the images in a cluster into a joint BoVW representation and perform inverted file retrieval to find the best matching scene models for a query image.

\PAR{Classification.}%
An alternative approach is to view the task as a classification problem where each landmark is a class. Gronat \etal \cite{Gronat13CVPR} learn exemplar SVMs based on the BoVWs of the visual features of the database images. Li \etal \cite{Li09ICCV} learn a multi-class SVM and additionally use the BoWs (bags-of-words) of the textual tags of the images as features. Bergamo \etal \cite{Bergamo13CVPR} use a similar approach, but perform classification using 1-vs-all SVMs. Instead of using approximate k-Means \cite{Philbin07CVPR} for feature quantization, they reconstruct the landmarks using structure-from-motion and train random forests on the descriptors of each structure-from-motion feature track. These random forests are then used for quantizing descriptors. While discriminative methods often yield higher accuracy than nearest neighbor matching, they also have disadvantages. For example, they assign \emph{every} image a landmark label regardless of whether it contains a landmark. Moreover, discriminative models need to be re-trained every time new images and landmarks are added.

\PAR{Pose Estimation.}%
The goal of pose estimation is to determine the camera location and orientation for a given query image. There are several approaches for solving this task by matching the query against street level imagery such as Google Street View panoramas \cite{Baatz10ECCV,Chen11CVPR,Knopp10ECCV,Torii11MVW,Schindler07CVPR,Johns14IJCV} using local feature based image retrieval \cite{Nister2006CVPR,Philbin07CVPR,Sivic03ICCV}. Other approaches are based on 3D point clouds created by applying structure-from-motion on Internet photo collections or manually collected photos  \cite{Sattler11ICCV,Sattler12ECCV,Li10ECCV,Li12ECCV}. Since image retrieval methods cannot be applied here, these approaches directly match the query descriptors against the descriptors of the image features that the 3D points were reconstructed from. After a set of 2D-3D correspondences has been established, the camera pose is determined by solving the perspective-n-point (PnP) problem \cite{Hartley04}. Since the descriptor matching problem becomes computationally expensive when matching against very large 3D models, hybrid methods have been proposed that \cite{Cao13CVPR,Irschara09CVPR,Sattler12BMVC} first perform efficient image retrieval using inverted files and then solve the PnP problem based on the relatively small set of 3D points associated with the 2D features of the retrieved images.

It is important to realize that camera pose does not necessarily
determine what is visible in the image. Even though the camera is in
front of a landmark, the landmark might not be visible due to
occlusion or the user might be taking a picture of a non-stationary
object or an event near that landmark. Moreover, pose estimation
relies on either regularly sampled images (\eg Google Street View
panoramas), which are not available everywhere, or
structure-from-motion reconstructions, which are not always possible
to compute robustly.

Because of the disadvantages of \emph{Classification} and \emph{Pose
  Estimation} based approaches, we focus on \emph{Image Retrieval}
based approaches in this evaluation.

\subsection{Eliminating Redundancy}
Several methods have been proposed to reduce the size of the visual
search index. An obvious method is to apply standard compression
techniques \cite{Jegou09ICCV}, which reduces memory consumption at the
cost of computational efficiency. 
Instead, we are interested in \emph{eliminating redundancy} already before index construction.

Several works have addressed this problem at the \emph{image level}, \ie by removing redundant images from the index.
Li \etal \cite{Li08ECCV} summarize the input image collection in a set
of \emph{iconic} images by applying k-means clustering based on Gist
descriptors, and use only these images to represent a landmark in
retrieval.
Gammeter \etal \cite{Gammeter10ECCV} identify sets of very
similar images using complete-link hierarchical agglomerative
clustering and replace them by just one image. This step
yields a slight compression of the index without loss in performance.
Instead of performing clustering Yang \etal \cite{Yang11ICME} only determine a set of canonical views by applying PageRank on the matching graph of the image collection. They then discard all other views and match the query only against the canonical views.

Other works have addressed the problem at the \emph{feature level}.
Turcot \etal \cite{Turcot09LAVD}
perform a full pairwise matching of the images in the dataset and remove all
features that are not at least once inliers \wrt a homography. They
report a significant reduction of the number of features while
maintaining similar retrieval performance.
Avrithis \etal \cite{Avrithis10MM} and Johns \etal \cite{Johns11ICCV}
combine the images in a cluster into a joint BoVW
representation. Avrithis \etal \cite{Avrithis10MM} use Kernel Vector
Quantization to cluster redundant features and keep only the cluster
centers. While this method only yields a slight compression, the
aggregation of features into a Scene Map brings significant
improvements in recognition performance.  Johns \etal
\cite{Johns11ICCV} performs structure-from-motion and summarize
features that are part of the same feature track.
Gammeter \etal \cite{Gammeter09ICCV} estimate bounding boxes around
the landmark in each image in a cluster and
remove every visual word from the index that never occurs inside a
bounding box. This is reported to yield an index size reduction of
about a third with decreasing precision.

There is also work in \emph{pose estimation} that aims to eliminate redundancy in the dataset. In their hybrid 2D-3D pose estimation approach, Irschara \etal \cite{Irschara09CVPR} generate a set of synthetic views by projecting the SfM points onto a set of virtual cameras placed at regular intervals in the scene. They then decimate the set of synthetic views using a greedy set cover approach that finds a minimal subset of views such that each view in the subset has at least 150 3D points in common with an original view.
Cao \etal \cite{Cao14CVPR} use a similar criterion, but instead of views, they decimate the set of points in an SfM point cloud used for localization. Instead of set cover, they use a probabilistic variant of the K-Cover algorithm.

\subsection{Semantic Annotation}
The most common approach to perform \emph{semantic annotation} of the
discovered landmark clusters is by statistical analysis of
user-provided image tags, titles and descriptions.
In order to remove uninformative tags like ``vacation'', Quack \etal
\cite{Quack08CIVR} first apply a stoplist and then perform frequent
itemset analysis to generate candidate names. These names are verified
by querying Wikipedia and matching images from retrieved articles
against the landmark cluster.
Zheng \etal \cite{Zheng09CVPR} also apply a stoplist and then simply
use the most frequent n-gram in the cluster.
Crandall \etal \cite{Crandall09WWW} deal with uninformative tags in a
more general way by dividing the number of occurrences of a tag in a
cluster by its total number of occurrences in the dataset.
Simon \etal \cite{Simon07ICCV} additionally account for tags that are
only used by individual users by computing a conditional probability
for a cluster given a tag, marginalizing out the users. 

Unfortunately, a much larger problem, also observed by Simon \etal
\cite{Simon07ICCV}, exists for the task of semantic assignment that is
much harder to fix: For most clusters accurate tags are simply not
available. In our analysis (Sec.~\ref{sec:semantics}), we will show for
which clusters these methods will still result in accurate
descriptions and point out the sources of this problem.

\section{Evaluation Setup}
\label{sec:evaluation}

\subsection{Dataset}
Our evaluation is based on the \textsc{Paris 500k} dataset \cite{Weyand10RMLE}
consisting of 500k images of the inner city of Paris collected from
Flickr and Panoramio. In contrast to many other datasets
\cite{Li08ECCV,Philbin07CVPR} the images were retrieved using a
geographic bounding box query rather than keyword queries to
ensure an unbiased distribution of touristic photos.

\subsection{Query set, Categories and Evaluation}
\begin{figure}[t]
  \centering%
  \includegraphics[width=.5\textwidth]{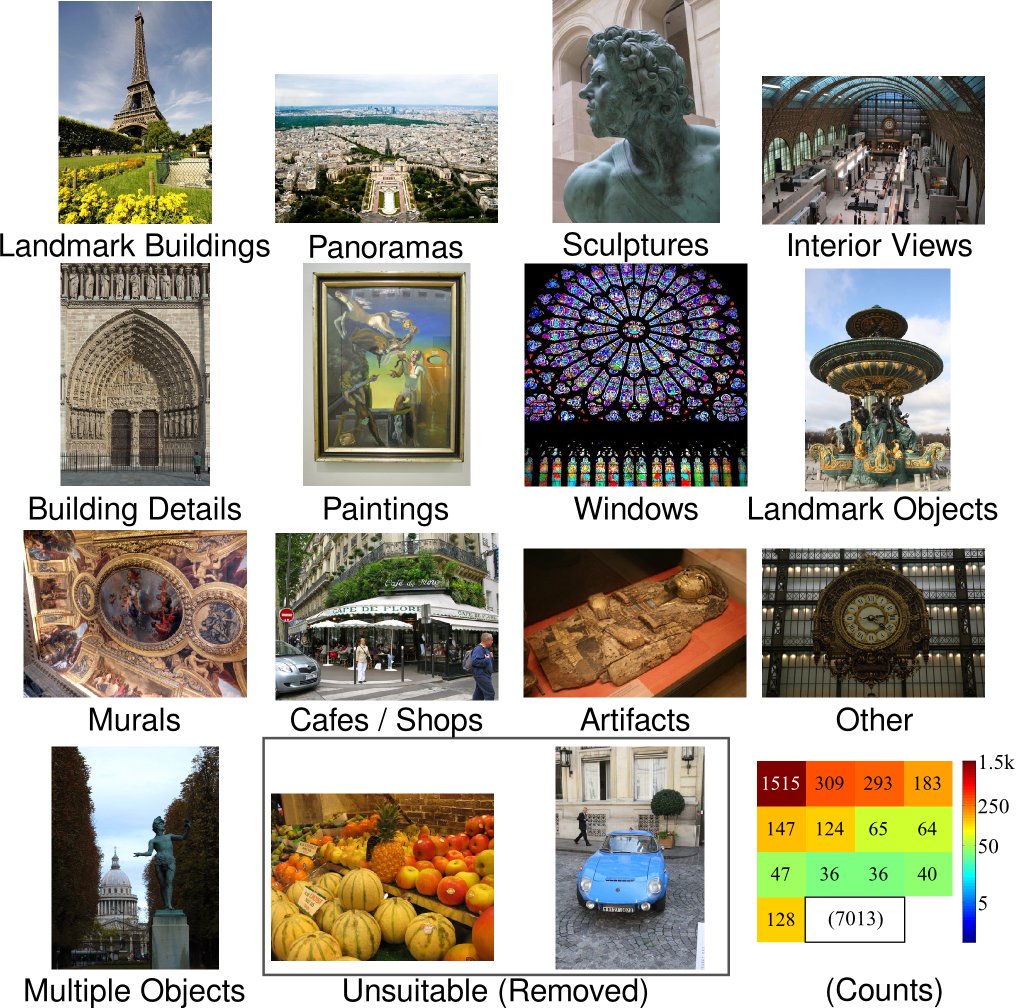}%
  \caption{Our set of query categories and the number of query images in each category (bottom right).}%
  \label{fig:query_categories}
\end{figure}

\begin{figure}
  \centering%
  \subfloat[]%
  {%
\includegraphics[width=.149\textwidth]{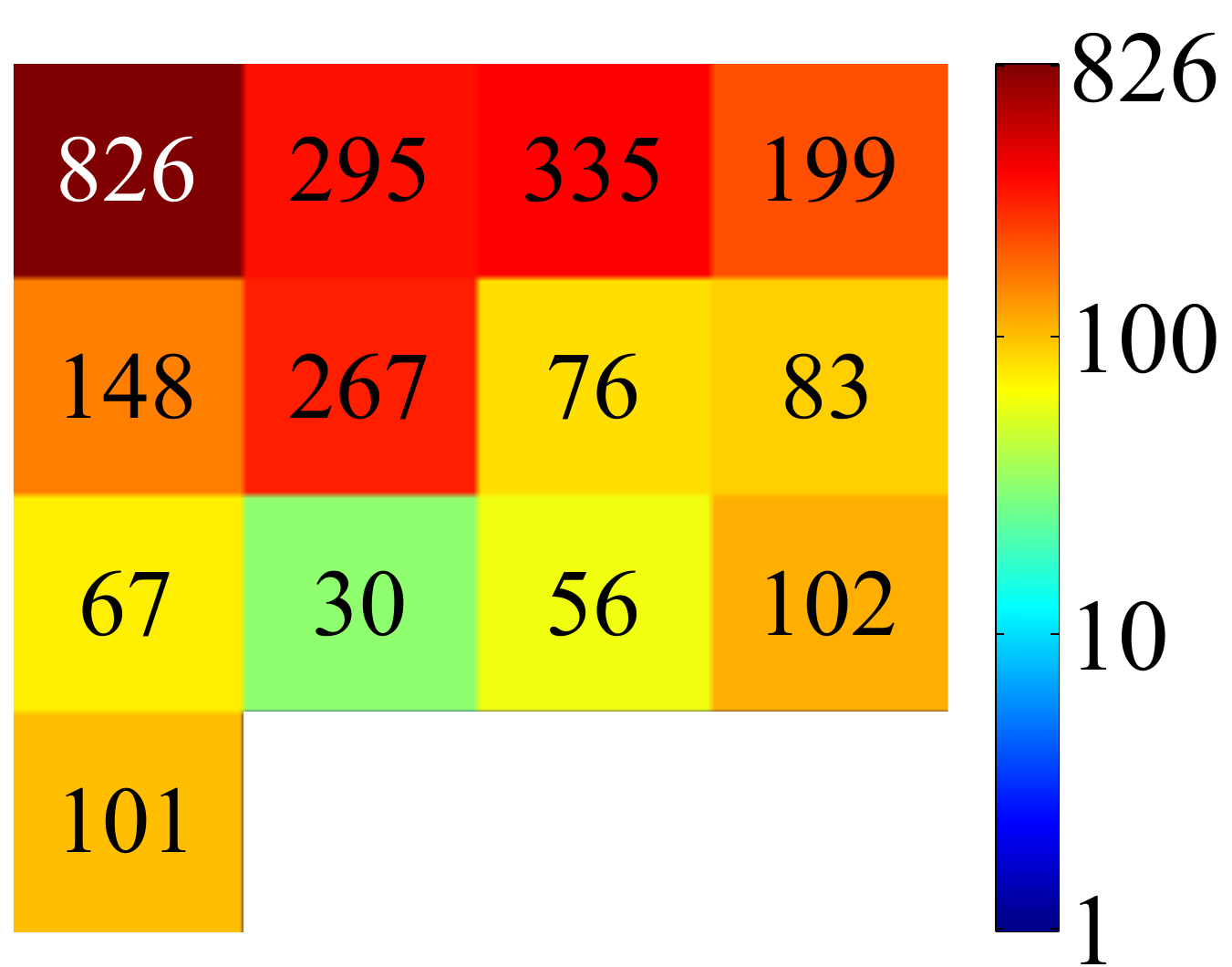}%
\label{fig:iconoid_colormatrices}%
}%
\hspace{5pt}%
  \subfloat[]%
  {%
\includegraphics[width=.149\textwidth]{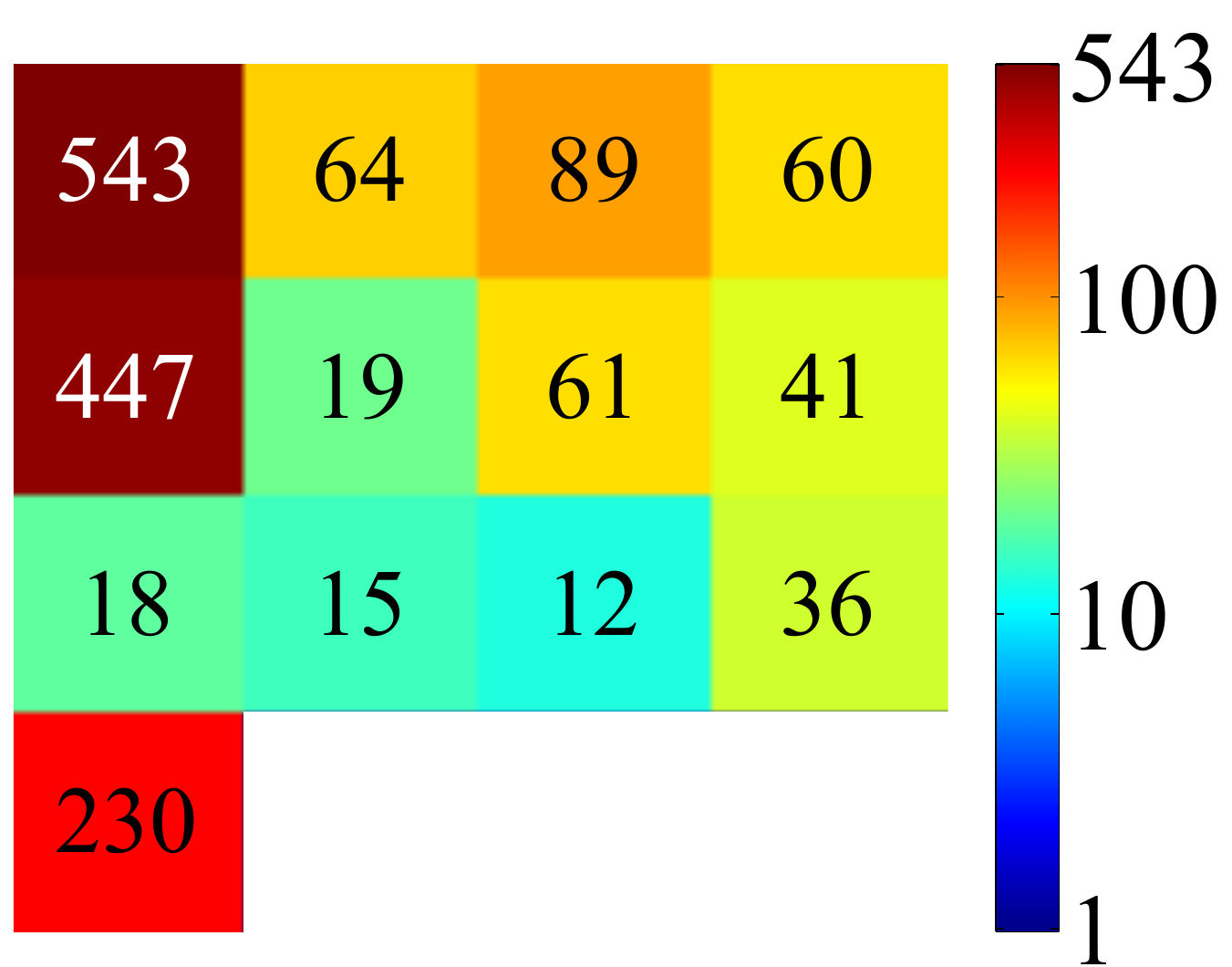}%
\label{fig:avg_cluster_size_per_category_colormatrix}%
}%
\hspace{5pt}%
  \subfloat[]%
  { \includegraphics[width=.149\textwidth]{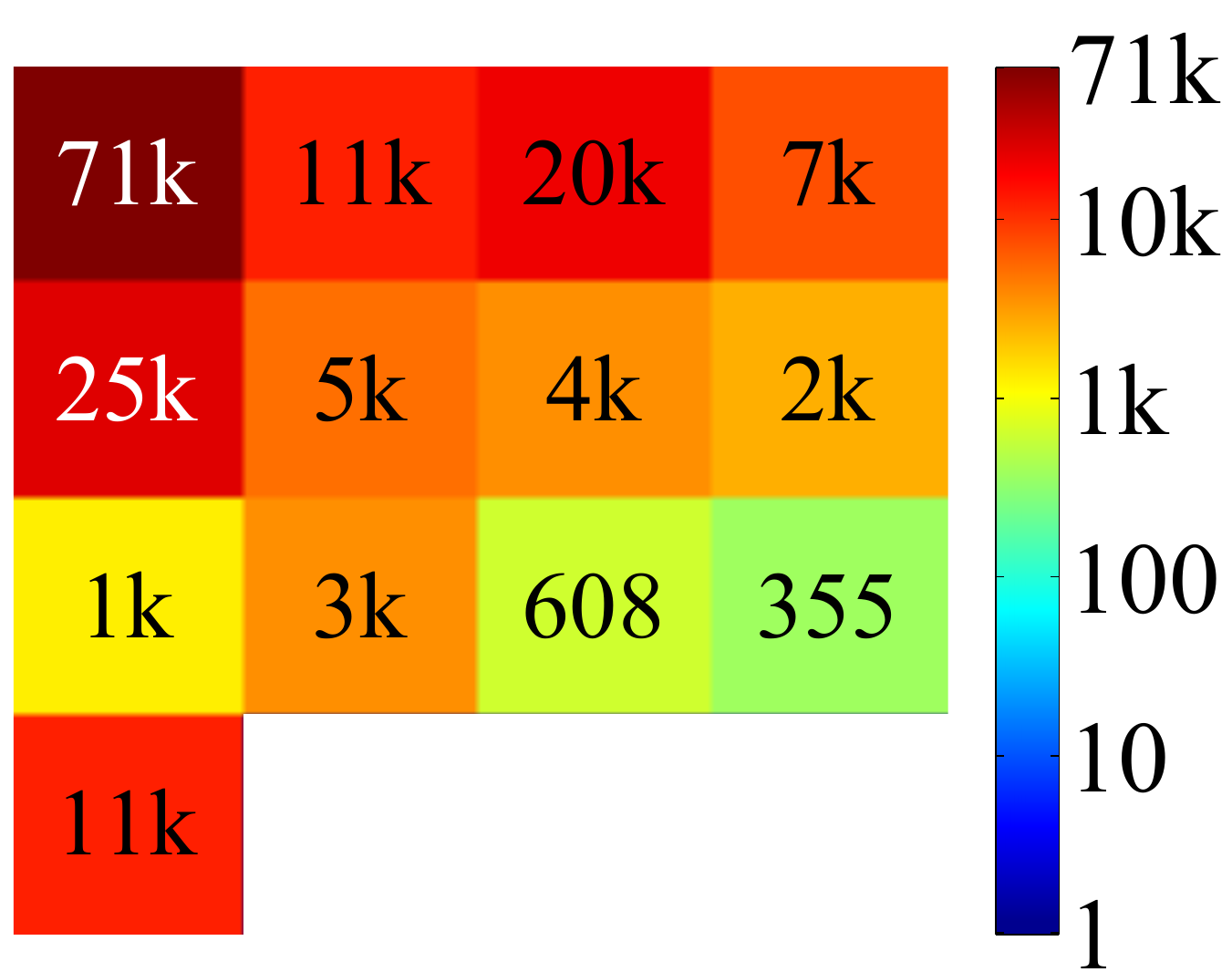}%
\label{fig:n_images_per_category_colormatrix}%
}%
  \caption{(a) Number of Iconoid clusters in each category. (b) Average cluster size. (c) Total number of images in clusters. Categories are in the same order as in Fig.~\ref{fig:query_categories}.}
\label{fig:discovered_categories_colormatrices}%
\end{figure}
To collect realistic queries for the task of automatic annotation of
photos uploaded to a photo sharing website, we downloaded 10k images
from the same geographic region as \textsc{Paris 500k} from Flickr and ensured
that they were no (near) duplicates of any image in the original
dataset.  Since we consider the task of \emph{specific object
  recognition}, not \emph{object categorization}, we filtered out
unsuitable queries like food, pets, plants or cars.  To only include
objects that have a \emph{chance} of being recognized based on the
\textsc{Paris 500k} dataset, we also exclude queries that do not match
\emph{any} image in \textsc{Paris 500k}, leaving 2,987 queries.
We manually grouped the queries into the categories of
Fig.~\ref{fig:query_categories} in order to enable an analysis of the
recognition performance for each query type. We summarize non-building
objects such as bridges, fountains or columns under the \emph{Landmark
  Objects} category. The \emph{Artifacts} category contains historic
objects such as sarcophagi or ancient tools. Objects that do not fit
into any other category were categorized as \emph{Other}. Note that
there is a large variance in the number of query images for each
category (given on the bottom right of
Fig.~\ref{fig:query_categories}).  The average scores over all query
images we provide in this paper therefore have a bias towards the
larger categories. This effect is desired, since we want the query
distribution to be representative of a real application in a photo
auto-annotation system. In addition to this, however, we will also
provide a detailed analysis for all 13 categories, focusing on four
categories representative of different use cases, namely
\emph{Landmark Buildings}, \emph{Paintings}, \emph{Building Details}
and \emph{Windows}.

\subsection{Image Retrieval}
Because it has become the de-facto standard in landmark recognition
\cite{Avrithis10MM,Gammeter09ICCV,Gammeter10ECCV,Kalantidis11MMTA}, we
use the vector space model for image retrieval
\cite{Sivic03ICCV,Philbin07CVPR}.  We extract SIFT \cite{Lowe04IJCV}
features from each image in the dataset and jointly cluster them using
approximate k-means \cite{Philbin07CVPR} to generate a dictionary of
1M visual words. We then vector-quantize the SIFT descriptors based on
this vocabulary and represent each image as a BoVW histogram. We build
an inverted file index from the BoVW histograms of the entire database
\cite{Sivic03ICCV} to enable efficient retrieval. Given a query image,
we extract its BoVW as described above, query the index and rank the
retrieved images by their $\mathit{tf}*\mathit{idf}$ scores \wrt the
query \cite{Sivic03ICCV}. For the top-300 matches we attempt fitting a
homography using SCRAMSAC \cite{Sattler09ICCV} and consider a match
verified if it has 15 or more inliers with the query. We then rank
verified images above unverified images \cite{Philbin07CVPR}.

\subsection{Scoring}
\label{subsec:scoring}
We would like to evaluate the performance of a landmark recognition
system in a realistic scenario. In a photo auto-annotation
application, the system should assign a user's photos reliable labels
without supervision. A mobile visual search app like Google Goggles
can also give the user a small selection of objects and let them pick
the correct one. Therefore, we consider only the top-3 objects
returned by the system.

For annotating recognition results, we showed the query and the iconic
image of the recognized object to raters and asked them to rate the
object's relevance to the query as ``good'' if it is the exact object
in the query image, ``ok'', if it is somewhat relevant to the query,
and ``bad'' if it is irrelevant. An object should be rated as ``ok'',
\eg if the query image shows a whole building, but the match only
shows a detail of that building, or vice versa. In case the query is a
detail of a building and the recognized object is a different detail
of the same building, the match should be rated as ``bad''. If the
query shows multiple landmarks, and the object is one of them, the
match should still be rated as ``good''.

Based on this rating, we define four scores:
\emph{good-1} is the fraction of queries with a ``good'' top-1 match;
\emph{ok-1} is the fraction of queries with an ``ok'' or ``good''
top-1 match; \emph{good-3} is the fraction of queries with a ``good''
match in the top-3; and \emph{ok-3} is the fraction of queries with an
``ok'' or ``good'' match in the top-3.

\subsection{Baseline Recognition Performance}
\begin{table}
  \centering%
\footnotesize
\begin{tabular}{lrrrr}
 Category            &   \%good-1  &   \%ok-1  &   \%good-3  &   \%ok-3  \\
\hline
 Landmark Buildings  &   94.32  &   98.22  &   97.62  &   98.42  \\
 Panoramas           &   87.70  &   95.15  &   91.59  &   95.79  \\
 Sculptures          &   92.15  &   95.56  &   94.20  &   96.25  \\
 Interior Views      &   85.25  &   89.07  &   89.07  &   92.35  \\
 Building Details    &   87.76  &   91.84  &   89.80  &   91.84  \\
 Paintings           &   97.58  &   98.39  &   98.39  &   98.39  \\
 Windows             &   95.38  &   95.38  &   95.38  &   95.38  \\
 Landmark Objects    &   93.75  &   96.88  &   96.88  &   96.88  \\
 Murals              &  100.00  &  100.00  &  100.00  &  100.00  \\
 Cafes / Shops       &   80.56  &   80.56  &   83.33  &   83.33  \\
 Artifacts           &   91.67  &   91.67  &   94.44  &   94.44  \\
 Other               &   92.50  &   97.50  &   97.50  &   97.50  \\
 Multiple Objects    &   98.44  &   98.44  &   98.44  &   98.44  \\
\hline
 Total               &   92.74  &   96.42  &   95.58  &   96.92  \\
\end{tabular}
  \caption{Performance of plain image retrieval using the full dataset.}
  \label{tab:baseline_retrieval}
\end{table}
For an estimate of the difficulty of the different query categories,
we perform image retrieval against the full \textsc{Paris 500k} dataset and
manually rate the relevance of the top-3 images for each query
according to the above scheme
(Tab.~\ref{tab:baseline_retrieval}). Note that these results only show
the relevance of \emph{retrieved images}, not \emph{recognized
  objects}, but can serve as upper bounds for the recognition
performance for each category. In total, the top-1 match was ``good''
for 92.74\% and at least ``ok'' for 96.42\% of the queries. Since
images that did not have a match in the database are not used in the
query set, the remaining 3.58\% had only false-positive matches in the
top-3.

\section{Landmark Object Discovery}
\label{sec:landmarks}

The first step of building a landmark recognition system is to cluster
the image collection into \emph{objects}
\cite{Chum10PAMI,Li08ECCV,Philbin08ICCVGIP,Quack08CIVR,Weyand11ICCV,Johns11ICCV,Crandall09WWW,Li09ICCV,Avrithis10MM,Zheng09CVPR,Ji12IJCV}. A
guiding question for our evaluation is: What object types can be
discovered by such a clustering? As we motivated in
Sec.~\ref{sec:relwork:lmdiscovery}, we choose Iconoid Shift
\cite{Weyand11ICCV} as our analysis tool to answer this question,
since it produces a set of overlapping clusters, which can represent
``overlapping'' objects, \eg, both the entire facade of Notre Dame and
individual statue groups on it. In addition, it has intuitive
parameters for controlling the granularity and number of discovered
clusters.

\subsection{Iconoid Shift}
Iconoid Shift \cite{Weyand11ICCV} is a mode finding algorithm based on
Medoid Shift \cite{Sheikh07CVPR}, designed for efficient discovery of
(potentially overlapping) object clusters in large image collections.
Starting from a seed image, the algorithm performs Medoid Shift in
image overlap space until it converges onto a discovered object, for
which it returns an iconic image (or \emph{Iconoid}) as well as a
support set containing all images having a certain minimum overlap
with the Iconoid. We take this support set as the cluster. Iconoid
Shift alternates between two steps: (i) Exploration of all images that
overlap with the selected image by recursive image retrieval. (ii)
Selection of the image that has the highest overlap with all the
explored images.  Empirically, the returned clusters often correspond
to distinctive views of individual objects or buildings, with the
Iconoid picking out the most central view. Like Mean Shift
\cite{Cheng95PAMI,Comaniciu02PAMI} and the object discovery approach of Chum \etal \cite{Chum10PAMI,Chum09CVPR}, Iconoid Shift is typically
initialized with a set of \emph{seed} images and is then run until
convergence for each seed. A larger number of seeds therefore results
in higher computational demands, but also causes more objects to be
discovered.

\subsection{Clusters Discovered per Category}
\label{sec:clusters_per_cat}
To analyze what objects can be discovered by visual
clustering, we run Iconoid Shift on the \textsc{Paris 500k} dataset. Following
\cite{Weyand11ICCV}, we choose a kernel bandwidth of $\beta=0.9$, meaning that an image needs to have at least $10\%$ overlap with an Iconoid to belong to its cluster. We perform several runs of the algorithm using different numbers of seed images selected randomly from \textsc{Paris 500k} in order to analyze the tradeoff between runtime and the number of objects discovered.

To examine what kinds of objects the algorithm finds, we categorize
all resulting Iconoid clusters of at least size 5 using the scheme
from
Fig.~\ref{fig:query_categories}. Fig.~\ref{fig:discovered_categories_colormatrices}
shows the number of discovered clusters for each category, their
average size and the number of images covered by clusters.
\emph{Landmark Buildings} are the largest category with 826 clusters
covering 71k images. The average cluster size of \emph{Building
  Detail} is surprisingly large, because some clusters include many
photos of the full facades due to our low choice of overlap threshold
for Iconoid Shift.  \emph{Painting} clusters are small on average,
while \emph{Windows} have fewer but larger clusters.
\begin{figure}[t]
  \centering
  \subfloat[] {
    \label{fig:iconoids_per_category_for_different_n_seeds_abs}
    \includegraphics[width=.65\linewidth]{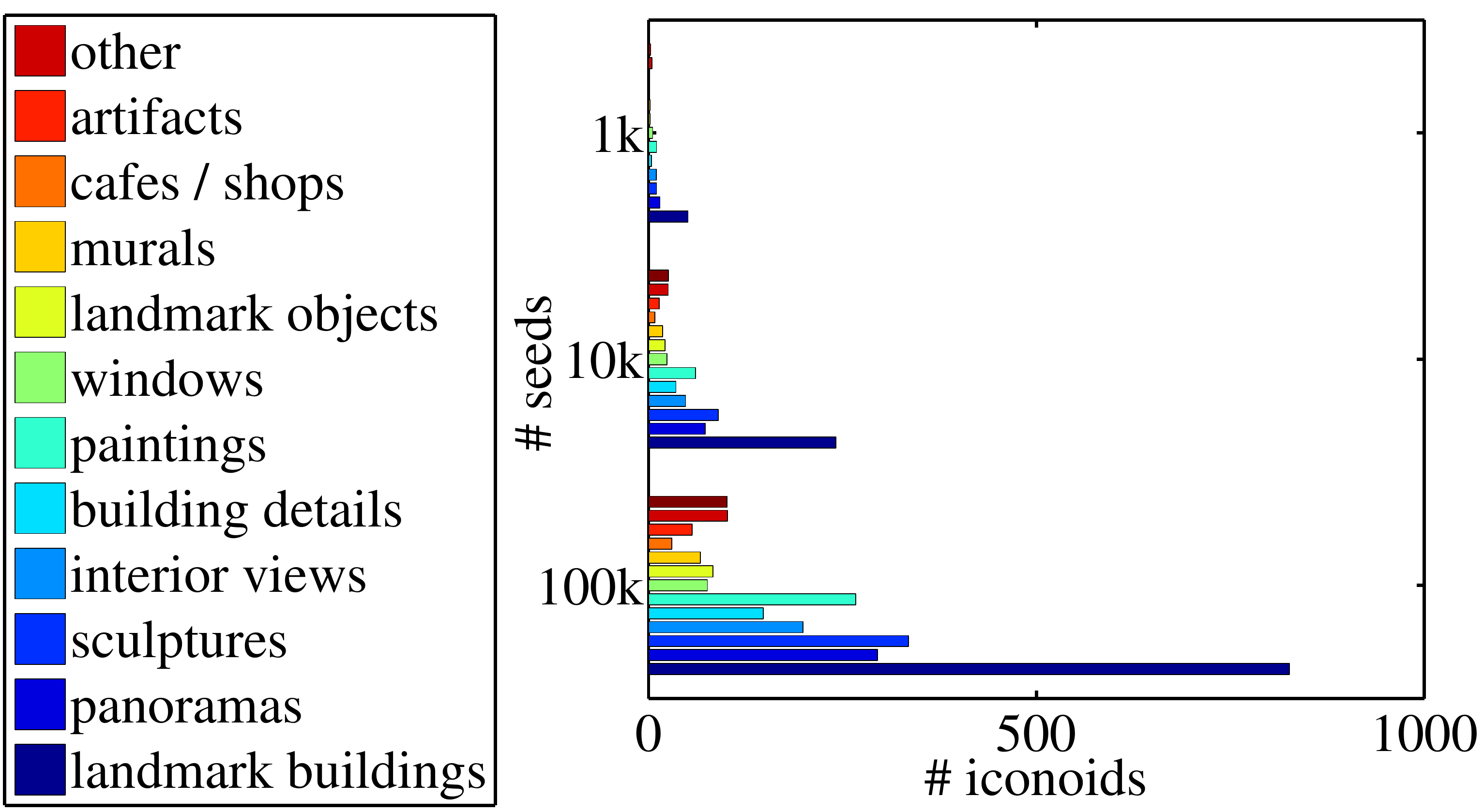}
  }
  \subfloat[] {
    \label{fig:iconoids_per_category_for_different_n_seeds_rel}
    \includegraphics[width=.35\linewidth]{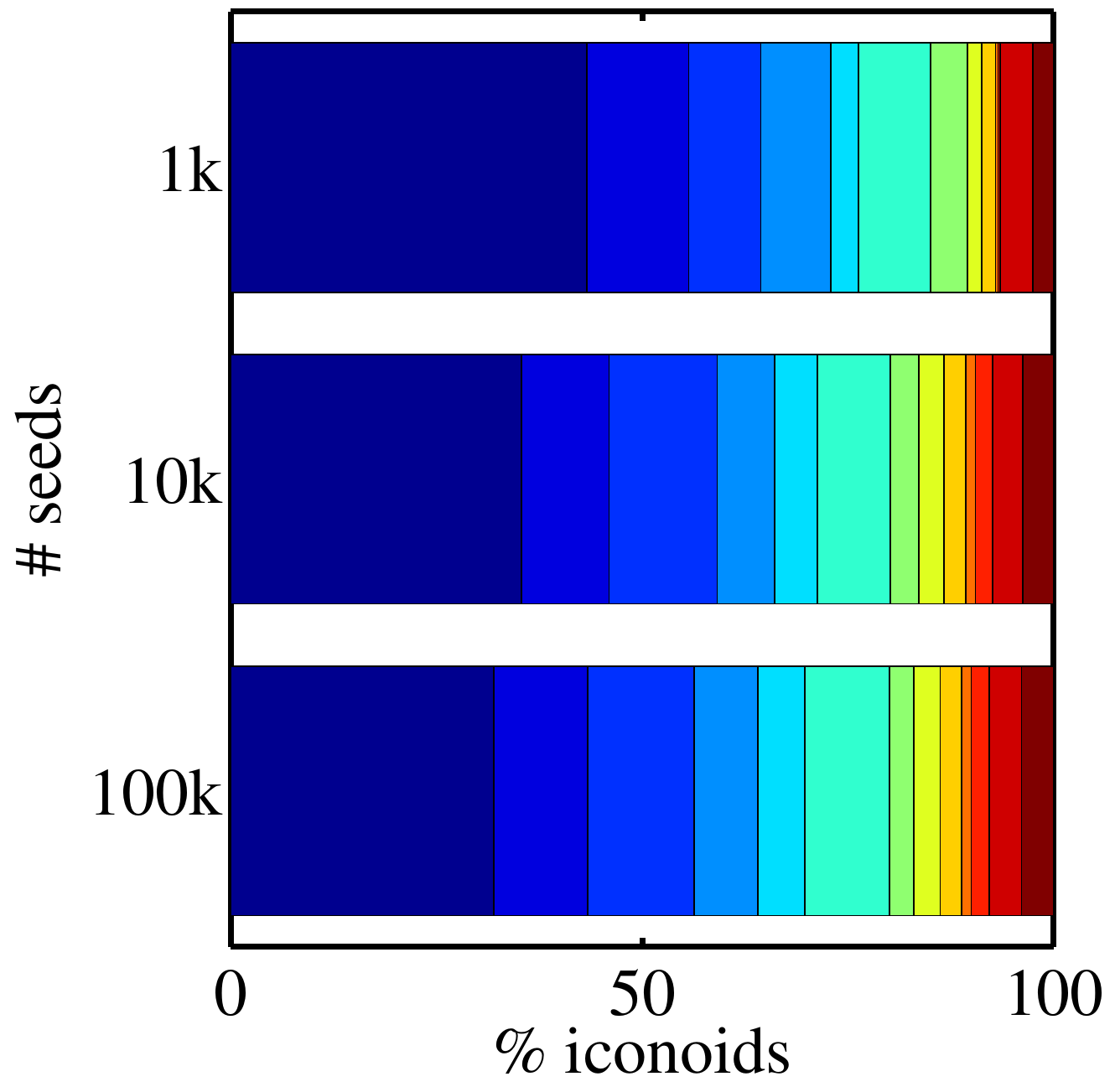}
  }
  \caption{Distribution of categories for different numbers of seeds counting only clusters of size 5 or larger. (a) absolute (b) relative.}
  \label{fig:iconoids_per_category_for_different_n_seeds}
\end{figure}

Fig.~\ref{fig:iconoids_per_category_for_different_n_seeds} shows the
effect of the number of seeds on the number of clusters discovered per
category.  When using only 1k or 10k seeds, the category distribution
remains relatively constant. The share of \emph{Landmark Building}
clusters decreases with increasing number of seeds
(Fig.~\ref{fig:iconoids_per_category_for_different_n_seeds_rel}),
since more of the smaller objects such as \emph{Paintings} or
\emph{Sculptures} are discovered.

\begin{figure}[t]
  \vspace{-8pt}
  \centering
  \subfloat[]
  {\label{fig:cluster_size_hist_over_n_seeds}
  \includegraphics[width=.33\linewidth]{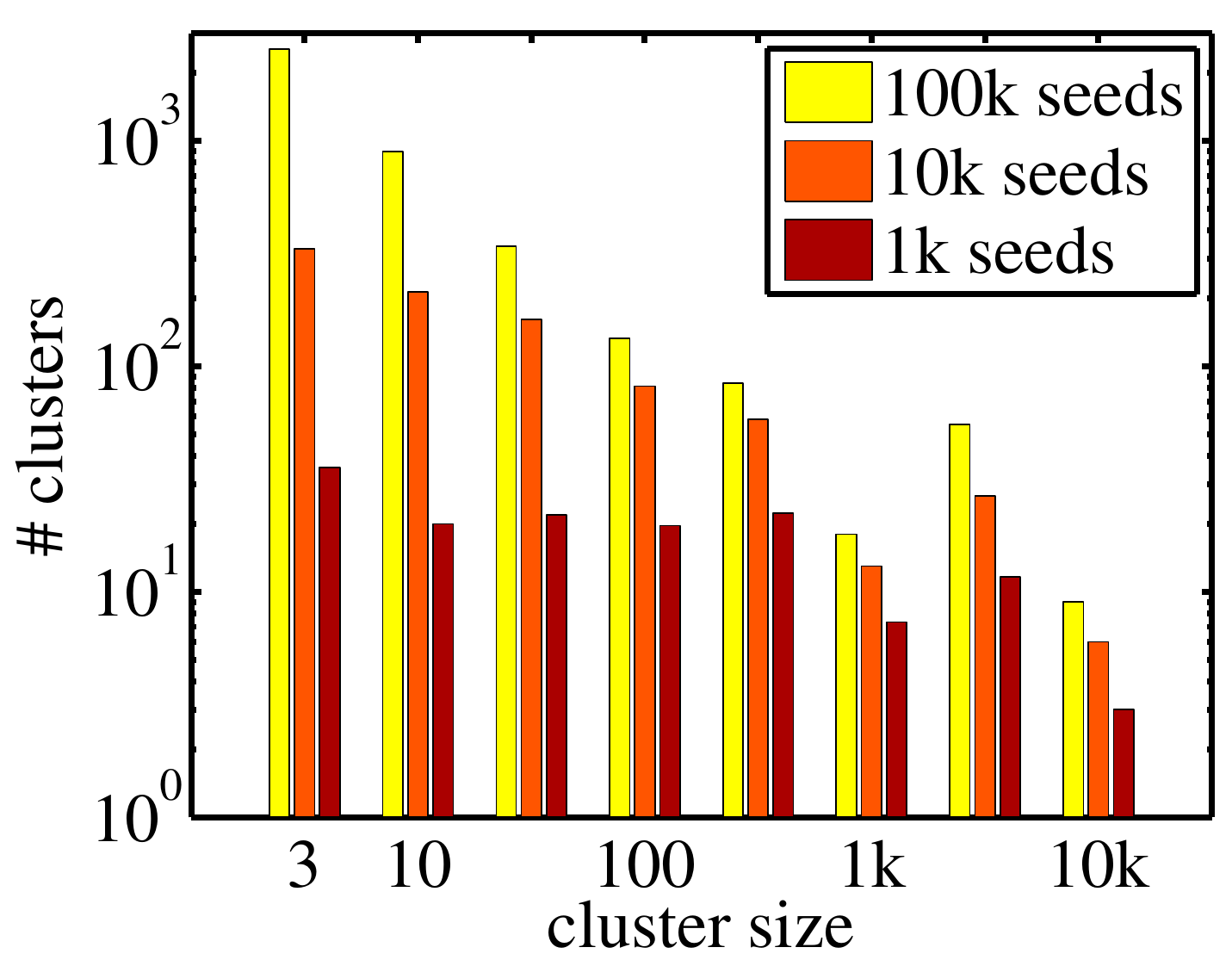}}
  \subfloat[]
  {\label{fig:iconoids_ge_10_over_seeds}
  \includegraphics[width=.33\linewidth]{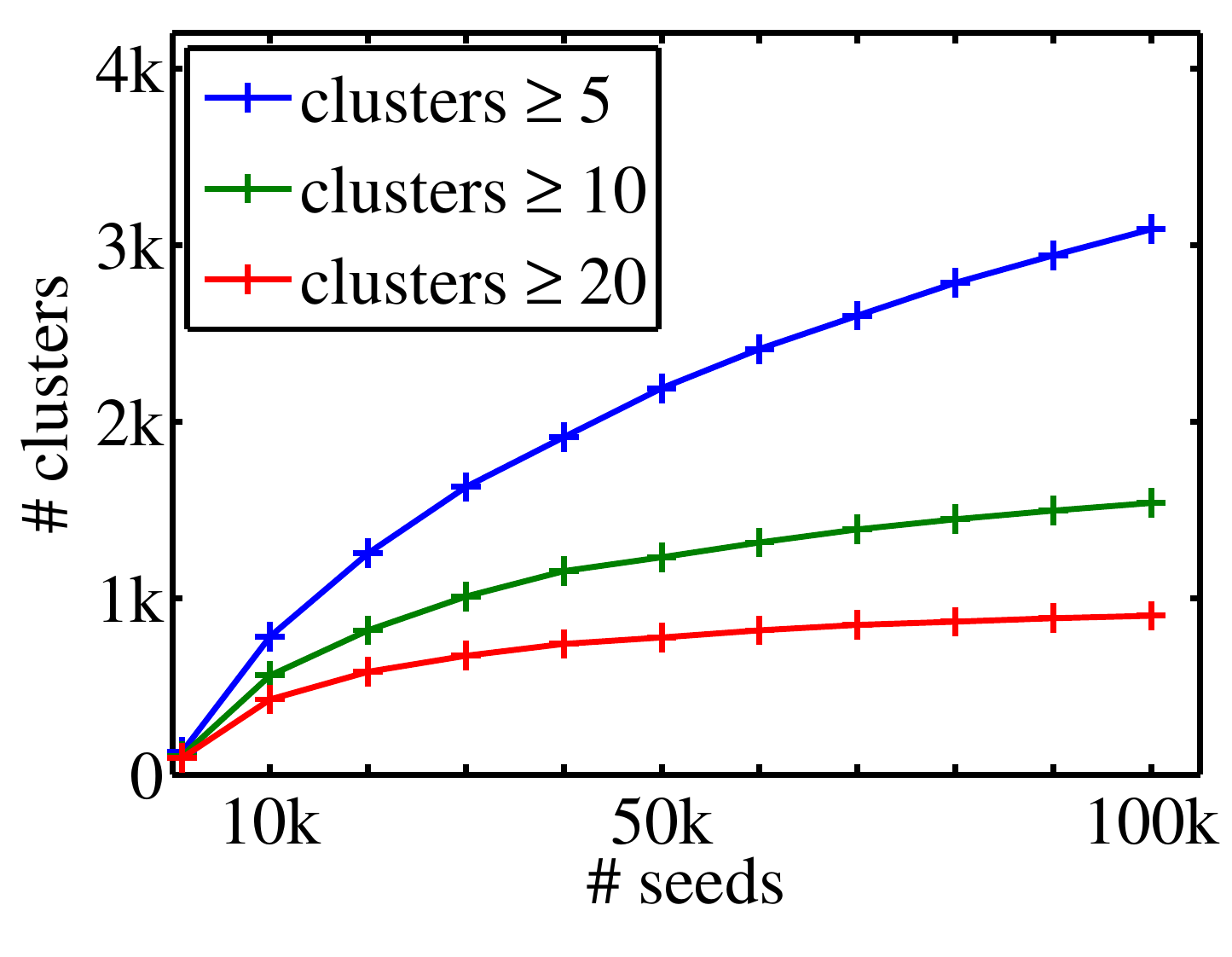}}
  \subfloat[]
  {\label{fig:images_covered_over_seeds} \includegraphics[width=.33\linewidth]{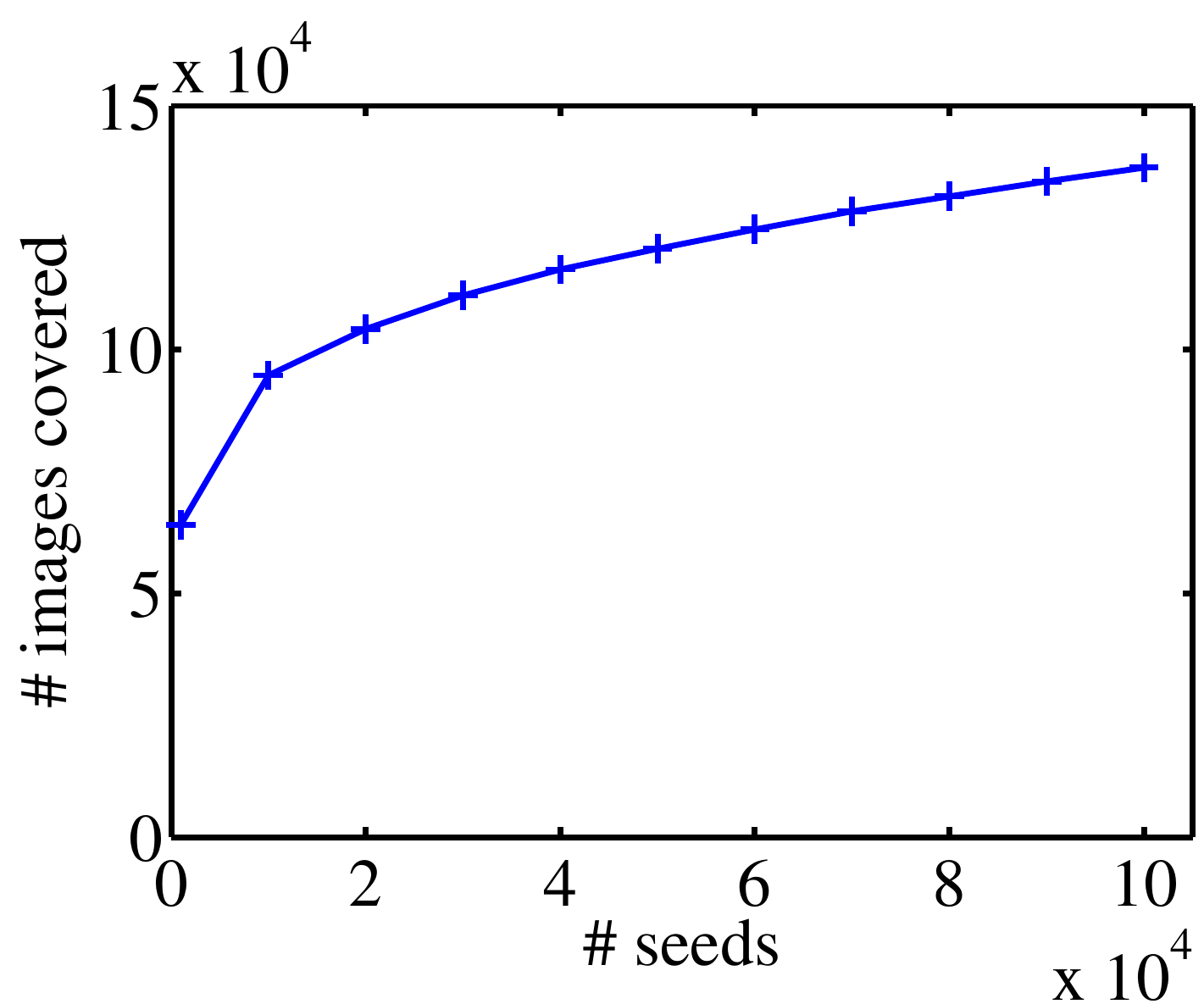}}
  \caption{(a) Cluster size distribution for different numbers of seeds (Note that both axes are logarithmic.) (b) Comparison of growth rates for different cluster sizes. (c) Total number of images covered by the clustering.}
  \label{fig:iconoid_sizes}
\end{figure}

\begin{figure}[t]
  \centering
  \includegraphics[width=.7\linewidth]{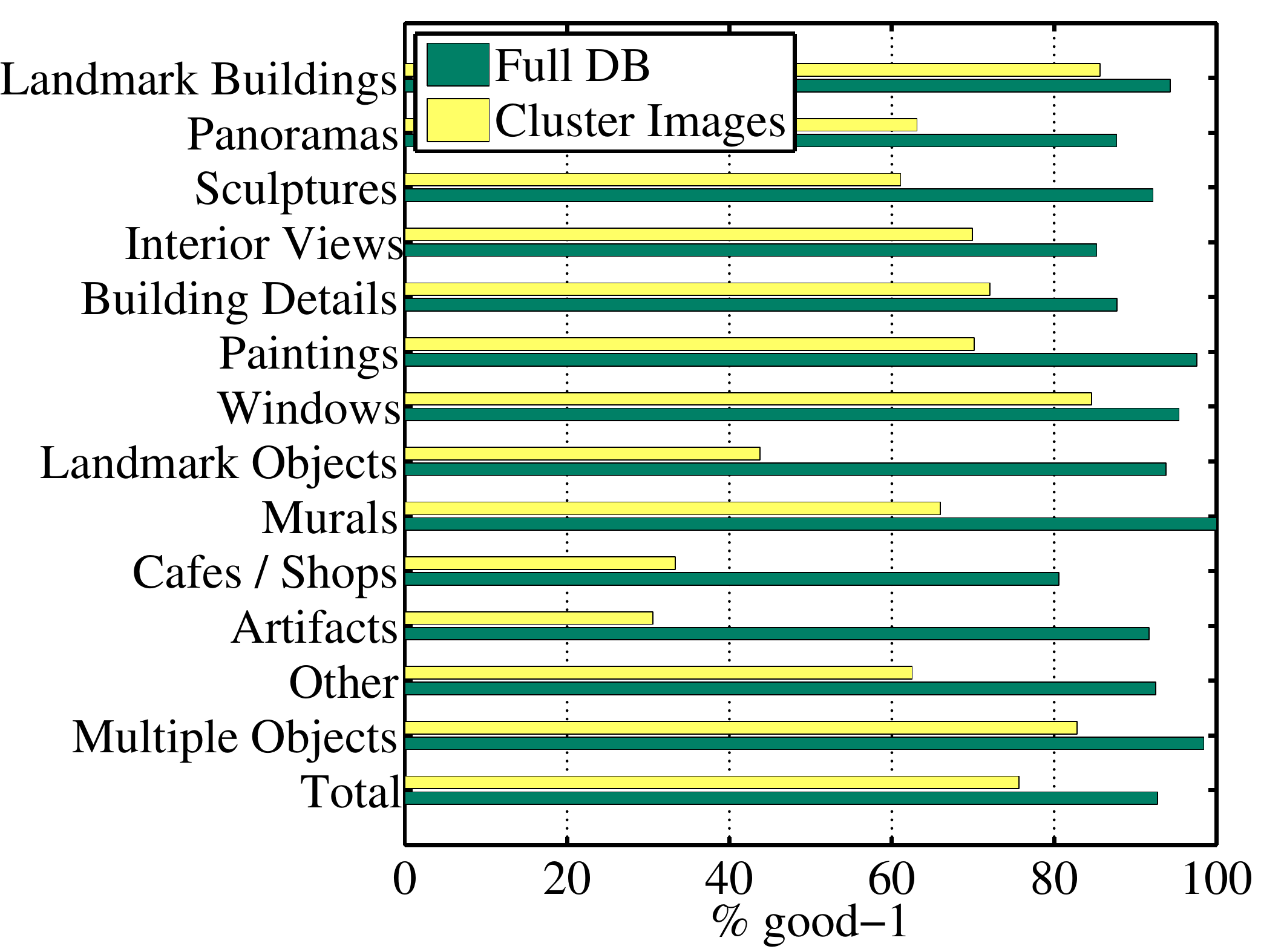}
  \caption{\emph{good-1} retrieval performance for the full database vs.\ only the images discovered by Iconoid Shift using 100k seeds.}
  \label{fig:baseline_retrieval_comparison}
\end{figure}

\subsection{Distribution of Cluster Sizes and Performance Gap}
The effect of the number of seeds on the distribution of cluster sizes
is shown in Fig.~\ref{fig:cluster_size_hist_over_n_seeds}. As reported
by \cite{Gammeter10ECCV}, the cluster sizes are power law
distributed. We can observe that the distribution shifts towards
smaller clusters when more seeds are used.
Fig.~\ref{fig:iconoids_ge_10_over_seeds} shows that the number of
large clusters flattens out more quickly than the number of small
clusters when increasing the number of seeds. This is because large
landmark clusters are found first, but more seeds help find more
obscure places and objects.  When using 100k seeds, 12,776 Iconoids
are returned, but only 3,088 of then contain 5 or more images.

Fig.~\ref{fig:images_covered_over_seeds} shows the total number of
database images covered for different numbers of seeds.  Using 100k
seeds, a total of 137,291 images (27.4\% of the dataset) are covered
by the clustering. The remaining images are either irrelevant or
missed by the clustering. To estimate how many and which objects the
clustering missed, we compare the retrieval performance using this
reduced set of images to the full set (\cf
Tab.~\ref{tab:baseline_retrieval}). The result
(Fig.~\ref{fig:baseline_retrieval_comparison}) shows in which
categories performance is lost and gives an upper bound on what a
landmark recognition system can achieve based on this
clustering. While for some categories, such as \emph{Landmark
  Buildings} or \emph{Windows}, the loss in performance is small,
other categories like \emph{Paintings}, \emph{Landmark Objects} or
\emph{Cafes} show a strong decrease, because their clusters are small
and thus more likely to be missed by the seeding process. This effect
is the main cause of the total performance gap of 17.05\% between the
full database and the cluster images.

\subsection{Discussion}
Since often-photographed objects are discovered first, seed-based
clustering can be computationally much more efficient than computing
the whole matching graph. To sufficiently cover seldom photographed
objects such as museum exhibits, a larger number of seeds is
necessary. The coverage of such objects could also be increased by
seeding strategies that avoid the bias to large clusters. Small object
discovery approaches \cite{Chum09CVPR,Letessier12ACMMM} or approaches
that crawl tourist guide websites \cite{Zheng09CVPR} might also help
cover these objects better and close the above performance gap
further. Whatever strategy is chosen, the results in
Fig.~\ref{fig:baseline_retrieval_comparison} show that such additional
steps are necessary if \emph{Landmark Objects}, \emph{Cafes / Shops}
or \emph{Artifacts} shall be recognized. For most of the following
experiments, we choose to use 100k seeds to ensure good coverage of
details and small objects.

\section{Landmark Object Recognition}
\label{sec:recognition}
By clustering a large collection of tourist photos, we have discovered
numerous interesting objects and determined a set of
representative images for each of them. To now recognize a new object
in a query image, a landmark recognition system performs
retrieval in the set of discovered object representatives
(Fig.~\ref{fig:retrieval_framework}). The open question here is: Given
a ranking of representatives, how to rank the objects they belong to
by their relevance to the query?  To this end, we compare five object
scoring methods and evaluate their respective tradeoffs of performance \vs database size and their suitability for different object categories.

\subsection{Ground Truth Generation}
\label{subsec:ground_truth}
For this evaluation, we introduce a new ground truth containing relevance ratings (Sec.~\ref{subsec:scoring}) of the Iconoids discovered with 100k seeds \wrt the query set.  An exhaustive relevance annotation of the 12,776 Iconoids for each of the 2,987 query images would require about 883 person-days of human work, assuming 2s of annotation effort per query-Iconoid pair. Therefore, we took two measures to reduce the amount of manual labor.
(1) We summarized queries showing exactly the same view into 2,042
groups, since the same Iconoids are relevant for them. For this, we
computed a pairwise matching of the queries and manually inspected
each pair of matching images, discarding all pairs that do not show
exactly the same view. We then constructed a matching graph from the
verified edges and computed its connected components. During
annotation, each group was represented by one image, and annotations
for it were transferred to all other members of the group.
(2) We automatically rated an Iconoid as irrelevant for a group of queries if \emph{none} of the Iconoid's representatives were spatially verified \emph{at least once} when querying an image retrieval system with \emph{each} query in the group.  To avoid false negatives in image retrieval, we performed an exact spatial verification by establishing correspondences using matching SIFT features that pass the SIFT ratio test \cite{Lowe04IJCV}. Since the landmark with the largest number of images in the \textsc{Paris 500k} dataset is the Eiffel Tower with about 20k images, any query can have at most 20k relevant images. To leave some room for ranking errors, we performed spatial verification for the top-30k retrieved images ranked \wrt their \textit{tf*idf} scores. The 26.8k remaining pairs of query groups and Iconoids were manually annotated according to the rating scheme introduced in Sec.~\ref{subsec:scoring}. Annotations were performed by 28 people over a period of 8 weeks. Each pair was shown to 3 people who were asked to rate it as \emph{good}, \emph{ok} or \emph{bad} according to the scoring scheme we introduced in Sec.~\ref{subsec:scoring}, and the final annotation was decided by majority voting. In the 1.8k cases where all three annotations were inconsistent, the image pair was passed to a fourth annotation for a definite annotation.

This ground truth, including the query images, the Iconoid clusters,
the query-Iconoid relevance annotations, and the query category
annotations is publicly available under
\footnote{\url{http://www.vision.rwth-aachen.de/data/paris500k/paris-dataset}}. We
believe that this ground truth will be useful to the landmark
recognition community, since (i) so far, there is no landmark
recognition dataset at this scale (3k queries, 13k clusters and 137k
representative images) (ii) this is the first dataset where the
clusters were produced by an actual landmark clustering algorithm
instead of keyword searches on Internet photo collections (iii) the
category annotations allow a detailed performance analysis for
different use cases, \eg, paintings or statues, while existing
datasets only focus on buildings.

\subsection{Methods}
\label{subsec:retrieval_methods}
We evaluate five object scoring methods:

\textbf{Center} represents objects only by the cluster center (\ie, Iconoid) of each
object and discards all other representatives, as done in Yang \etal
\cite{Yang11ICME}. The object ranking is then simply the same as the
representative ranking.

\textbf{Size} returns all objects with at least one matching
representative, and scores them by their cluster size, \ie, by the number of times they were photographed.

\textbf{Voting} lets each matching representative cast a vote for each object it
belongs to (note that clusters can overlap). Objects are then ranked by their
number of votes.

\textbf{Best Match} returns the object with the highest scoring representative, as done in, \eg, \cite{Quack08CIVR,Zheng09CVPR,Gammeter10ECCV,Gammeter09ICCV}. A difference in our case is that we are using a soft clustering, so a representative can belong to multiple objects. In this case, we return the object with the largest cluster size. This method can therefore also be viewed as a variant of the \emph{Size} method that only uses the best matching representative.

\textbf{Overlap} uses \emph{Homography Overlap Propagation}
\cite{Weyand11ICCV} to compute the overlap of the query with each
Iconoid. The method first computes the overlap region of the query
with the matching representative and then propagating this region into
the Iconoid via the shortest path in the Iconoid cluster's matching
graph. This is done using the homography overlap propagation (HoP)
algorithm from \cite{Weyand11ICCV}. If a query matches multiple
representatives of the same Iconoid, the overlap is computed for each
of them and the largest overlap is used. Objects are then ranked in
decreasing order of their Iconoids' overlaps with the query.

Note that, while \emph{Center} and \emph{Best Match} evaluate
strategies used in the literature, \emph{Overlap} is a novel strategy.
We now first compare the above methods using the objects discovered
with 100k Iconoid Shift seeds and then show the effect of the number
of seeds on their performance.

\subsection{Results}
\begin{figure}[t]
  \centering
  \includegraphics[width=\linewidth]{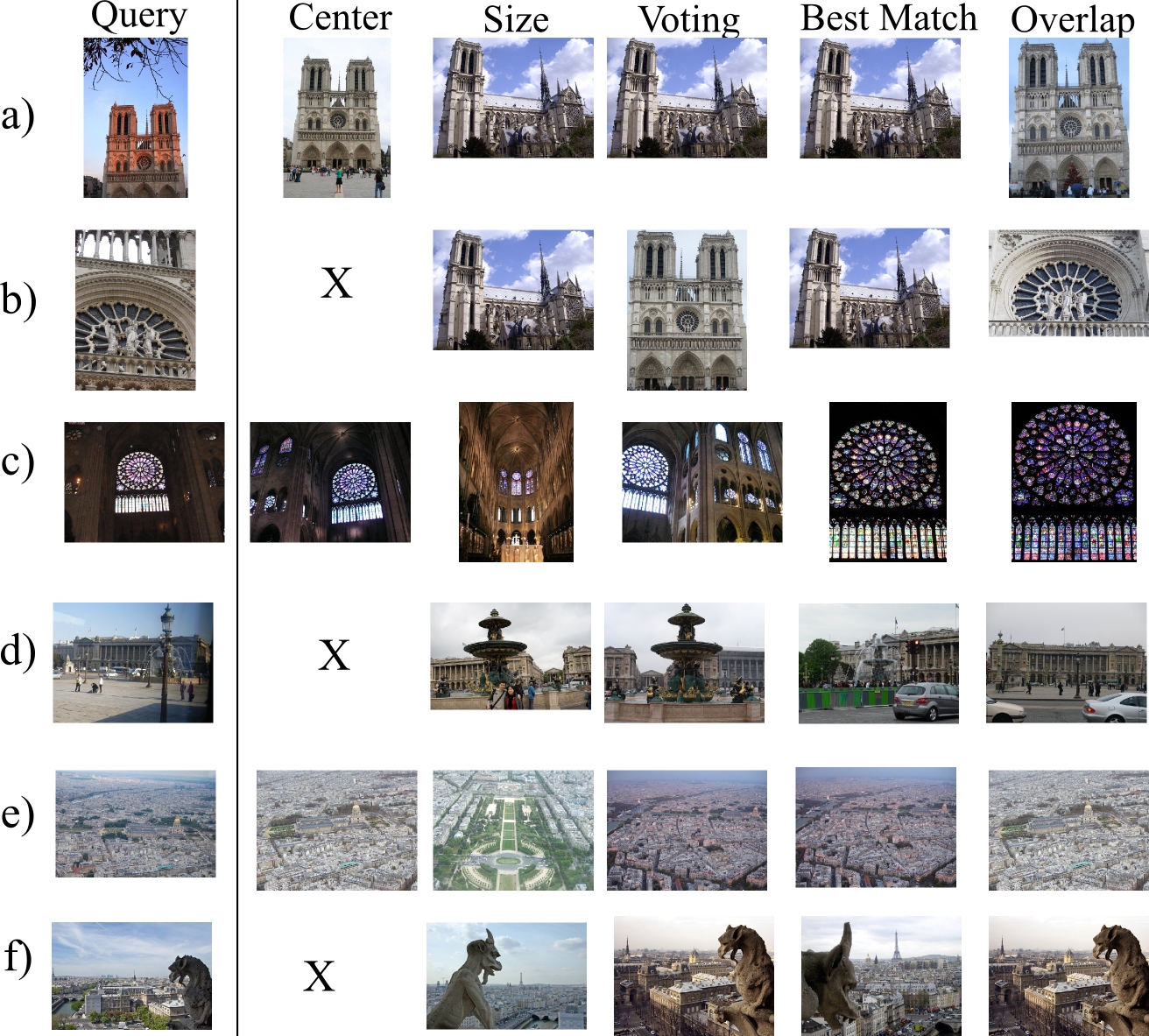}
  \caption{Top-scoring objects for different object scoring methods. ``X'' means that no objects were recognized.}
  \label{fig:retrieval_methods_visual_comparison}
\end{figure}
We evaluated the five approaches based on the ground truth introduced
in Sec.~\ref{subsec:ground_truth}. To estimate the error introduced by
the SIFT matching pre-filter we used to reduce annotation effort
(Sec.~\ref{subsec:ground_truth}), we performed a small control
experiment. We manually rated the relevance of the top-3 Iconoids
retrieved for each query using the \emph{Voting} method
(Sec.~\ref{subsec:retrieval_methods}) and found that 0.7\% of the
Iconoids rated ``good'' or ``ok'' were filtered out.  We believe this
small false-negative rate is still acceptable, since the simplified
annotation procedure significantly reduced the amount of manual labor.

The recognition performance of the different methods is compared
in Tab.~\ref{tab:retrieval_methods_quality}.
Fig.~\ref{fig:retrieval_methods_visual_comparison} shows the top
scoring objects for typical queries.
\emph{Center} finds images closely resembling the query, but often
fails to find \emph{any} matching objects, because the cluster centers are
not sufficient to recognize all objects under different viewing
conditions due to the limited invariance of the matching process.
However, since it only requires one image per object, it is by far
the fastest and most memory efficient method.
Because \emph{Size} chooses the largest cluster, it
often finds a viewpoint more popular than the query (popular
landmarks are often represented by multiple clusters from different viewpoints). Sometimes this
effect is desired since the largest cluster usually depicts the
object best, but it can also cause drift (Fig.~\ref{fig:retrieval_methods_visual_comparison}b-f), \ie instead of the query object, a nearby object is recognized.
\emph{Voting} finds the clusters with the most matching representatives. This makes it less prone to drift (Fig.~\ref{fig:retrieval_methods_visual_comparison}b,e,f) and causes it to achieve higher performance than \emph{Size}.
Despite being simpler than \emph{Size}, \emph{Best Match} outperforms
it. The reason is that \emph{Best Match} considers only the closest
matching representative to the query, making it less prone to drift
than \emph{Size} that also looks at farther away matches
(Fig.~\ref{fig:retrieval_methods_visual_comparison}c,e).
\emph{Overlap} has the best \emph{good-1} performance, because it computes the actual overlap of the query image with each cluster's iconic image and selects the iconic whose view is closest to the query (Fig.~\ref{fig:retrieval_methods_visual_comparison}a-f).

\begin{table}[t]
\footnotesize
  \centering
  \begin{tabular}{rrrrr}
    & \% good-1 & \% ok-1 & \% good-3 & \% ok-3 \\
    \hline
    Centers    & 39.60          & 45.56          & 42.99          & 46.03\\
    Size       & 57.11          & 73.32          & 66.42          & 76.26\\
    Voting     & 59.42          & \textbf{76.00} & 69.80          & \textbf{77.64} \\
    Best Match & 60.40          & 75.39          & 67.26          & 77.10\\
    Overlap    & \textbf{63.71} & 75.93          & \textbf{71.78} & 77.13\\
  \end{tabular}
  \caption{Performance of different object scoring methods.}
  \label{tab:retrieval_methods_quality}
\end{table}

\begin{figure}[t]
  \centering
  \includegraphics[width=\linewidth]{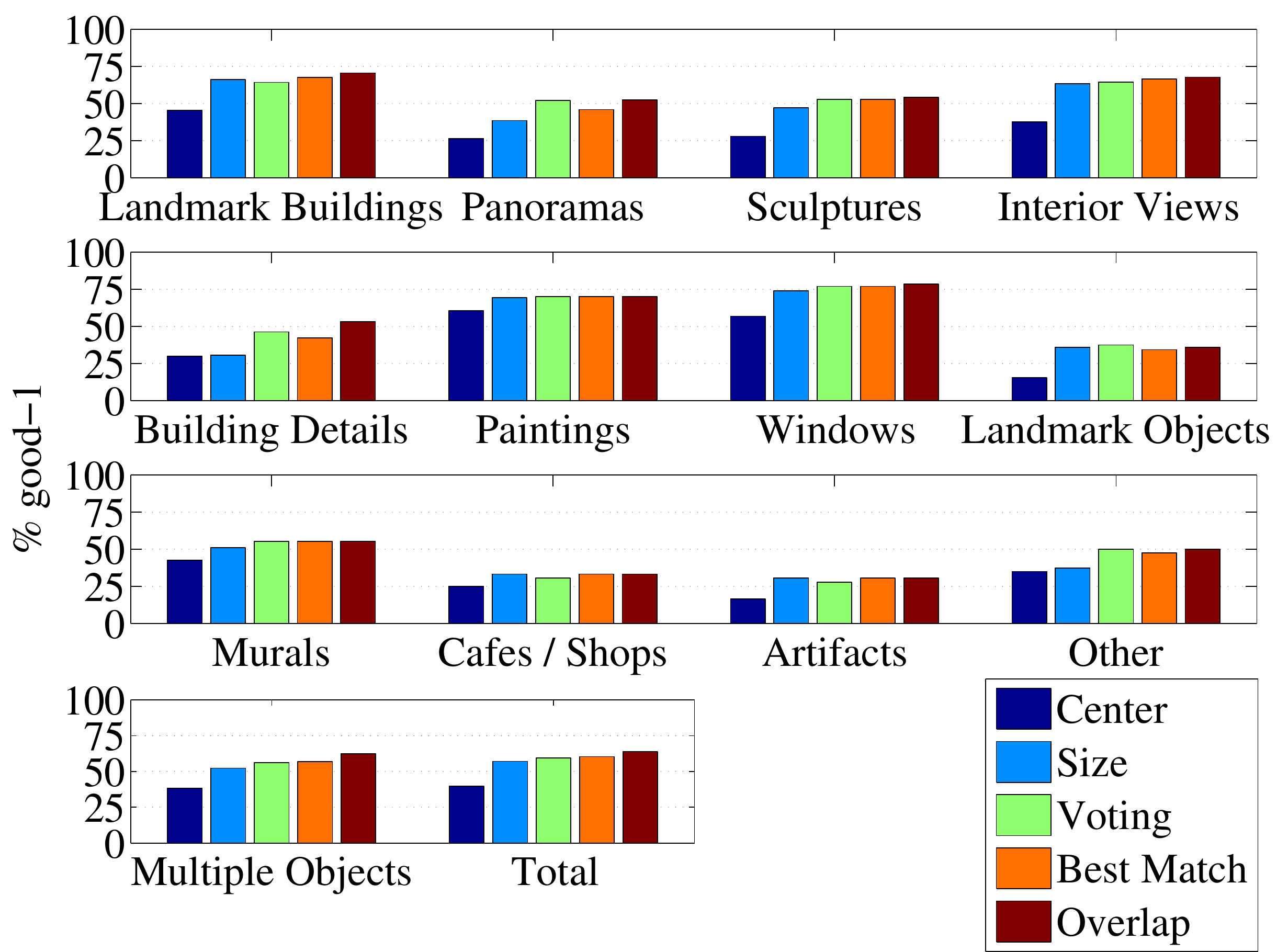}
  \caption{Performance of object scoring methods by query type.} 
  \label{fig:retrieval_methods_landmark_types}
\end{figure}

The \emph{good-1} performances by query type are shown in Fig.~\ref{fig:retrieval_methods_landmark_types}. \emph{Size} compares well to the other methods on \emph{Landmark Buildings} and \emph{Cafes / Shops}, where the largest cluster is often the correct one (\eg, the full view of a facade). On \emph{Paintings}, all methods including \emph{Center} have similar performance, because \emph{Paintings} are flat objects and are usually photographed under the same viewing conditions. Since this makes painting retrieval very easy, multiple representatives do not bring an advantage, explaining the relatively good performance of \emph{Center}. \emph{Size} performs worse than \emph{Voting} and \emph{Best Match} on \emph{Panoramas} (Fig.~\ref{fig:retrieval_methods_visual_comparison}e), because it tends to drift to more popular nearby views, moving the query object out of the field of view. The same effect occurs on \emph{Building Details} (Fig.~\ref{fig:retrieval_methods_visual_comparison}b), where \emph{Size} tends to return views of the whole building instead.
\emph{Overlap} has a particular advantage on classes where other methods tend to drift (\emph{Panoramas}, \emph{Building Details}, \emph{Multiple Objects}), since it usually finds the iconic image that best matches the photographed part (Fig.~\ref{fig:retrieval_methods_visual_comparison}a-f). \emph{Center} works relatively well for \emph{Windows} and \emph{Murals}, because, like \emph{Paintings}, they are only photographed from a limited range of viewing angles, making them easy to match.

\subsection{Effect of the Number of Seeds}
\begin{figure}[t]
  \centering
  \subfloat[]
  {\label{fig:iconoid_shift_seeds_retrieval_methods}
    \includegraphics[width=.375\linewidth]{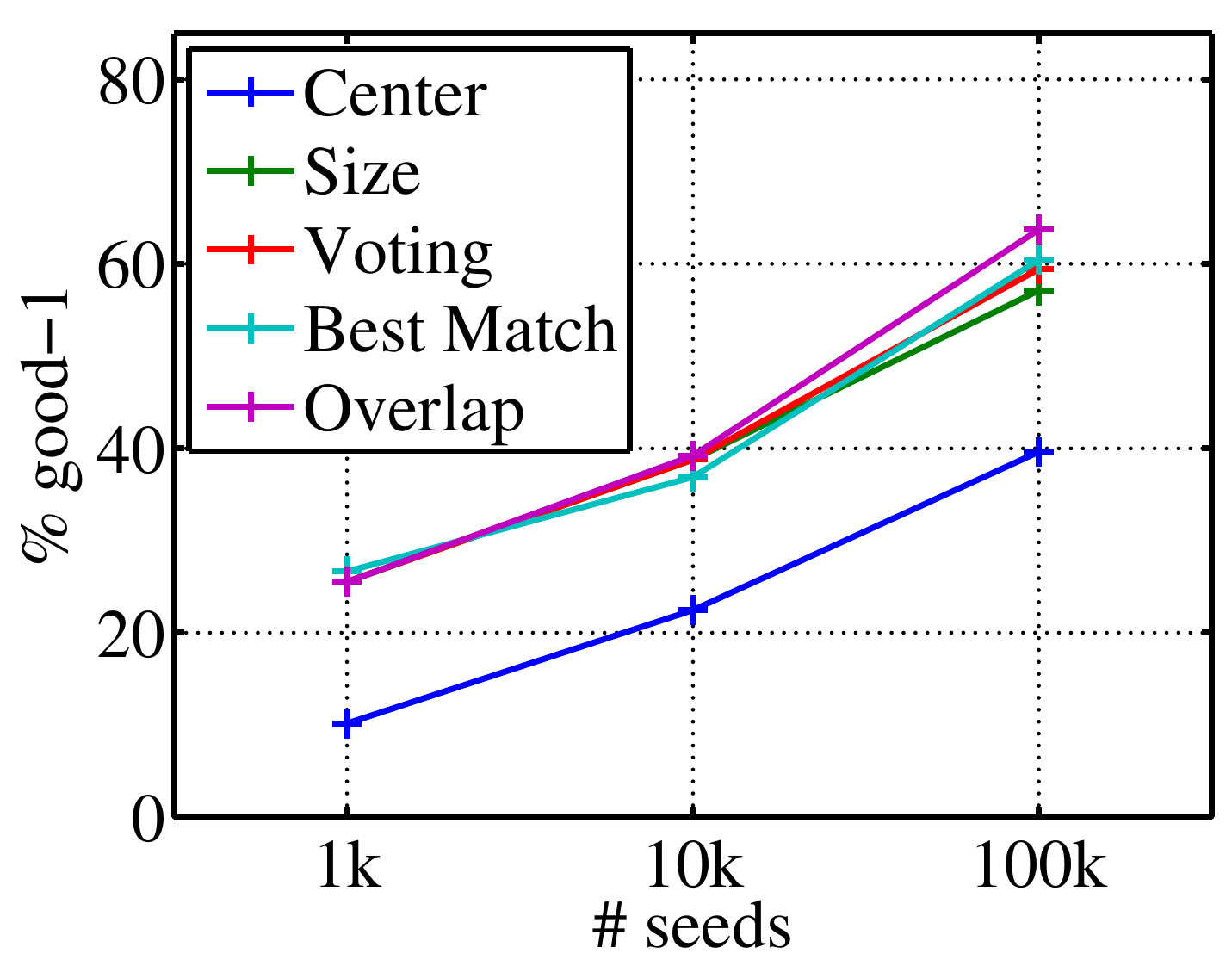}}
  \subfloat[]
  {\label{fig:iconoidshift_seeds_query_types}
    \includegraphics[width=.625\linewidth]{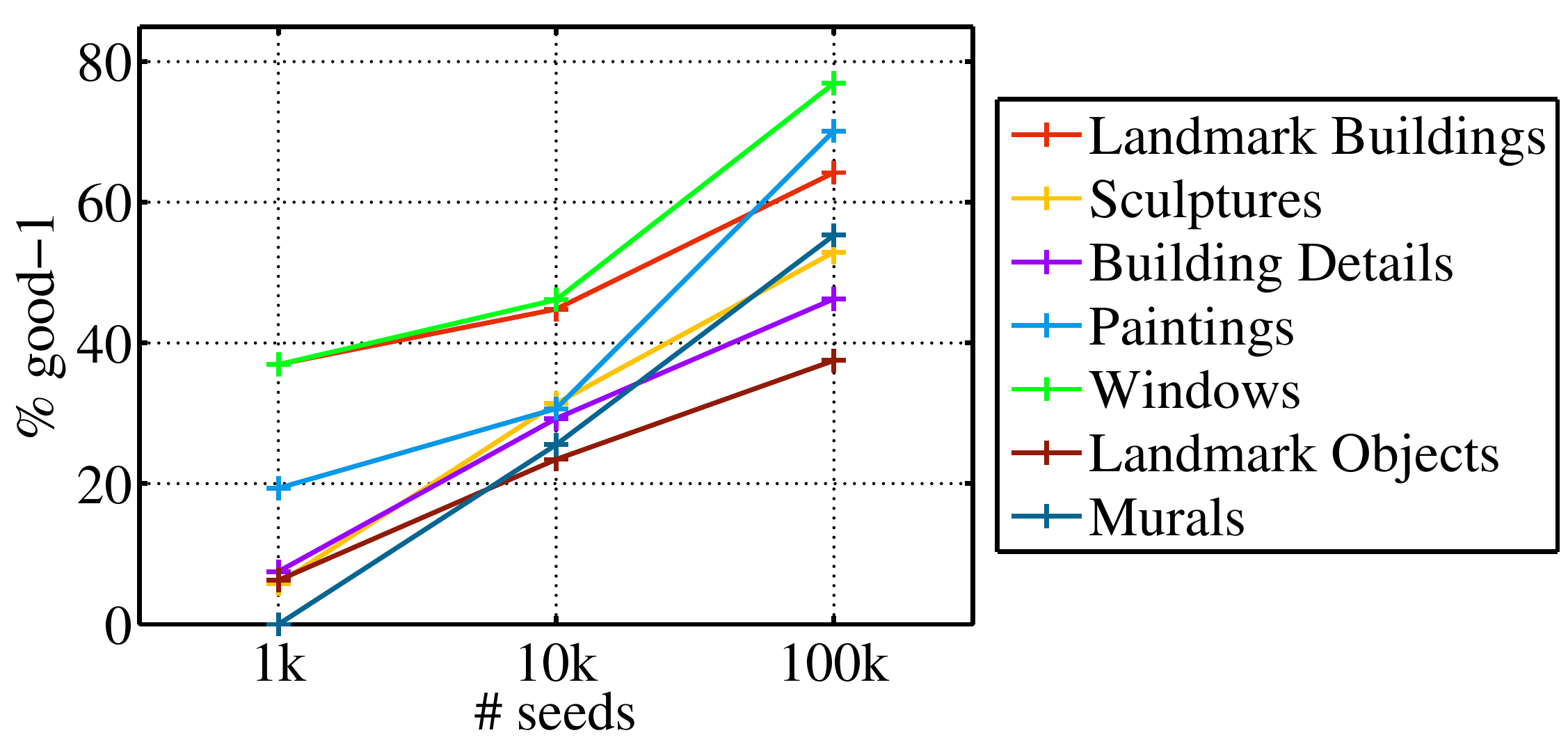}
  }
  \caption{(a) \textit{good-1} performance of object scoring methods for different numbers of Iconoid Shift seeds. (b) \textit{good-1} performance by query type using the \emph{Voting} method.}
\end{figure}
We now analyze the effect of the number of object clusters (which depends on the number of seeds) on the
performance of the five methods and compare object retrieval
performance by category.  As
Fig.~\ref{fig:iconoid_shift_seeds_retrieval_methods} shows, the
performance of the methods does not differ much for 1k and 10k seeds, except that
\emph{Center} consistently performs worst for the reasons explained above. The differences become slightly more pronounced at 100k seeds, where the density of objects is very high and methods that are less prone to drift to nearby objects gain an advantage. Conversely, this shows that simpler methods are sufficient if the object density is low.
Fig.~\ref{fig:iconoidshift_seeds_query_types} shows the performance
for different query types when using the \emph{Voting} method. \emph{Sculptures},
\emph{Paintings}, \emph{Windows} and \emph{Murals} show the steepest
improvement since they require more seeds to be sufficiently covered
by the clustering, while \emph{Landmark Buildings} can already be recognized when using a smaller number of seeds since they form large clusters that are discovered early. Surprisingly, \emph{Windows} have the highest recognition rate overall. The reasons for this are that (i) they are easy to recognize since they are flat, highly textured objects, and (ii) they get discovered already with few seeds, since \emph{Window} clusters are almost three times the size of \emph{Painting} clusters (Fig.~\ref{fig:avg_cluster_size_per_category_colormatrix}).

\subsection{Discussion}
The ideal choice of method varies by application: \emph{Center}
provides high performance for flat objects at low computational and
memory cost. \emph{Voting} has high accuracy across all
categories and its speed makes it applicable, \eg, for mobile visual
search. The popular \emph{Best Match} method has similar performance and efficiency, but is outperformed by
\emph{Overlap}, which has the highest performance overall but also the
highest computational cost. \emph{Overlap} is therefore better suited for offline
applications requiring high accuracy, \eg, photo auto-annotation.

Even when looking at the best performing method, \emph{Overlap}, there is still a difference of 11.98 percent points between the \emph{good-1} performance of object retrieval (63.71\%) and the \emph{good-1} performance of 75.69\% of plain image retrieval (Fig.~\ref{fig:baseline_retrieval_comparison}). The cause for this \emph{clustering gap} could be either a too coarse clustering or imprecise object ranking. If we consider that the difference between \emph{good-1} and \emph{good-3} for \emph{Overlap} (8.07\%) is larger than the gap of 3.91 percent points between the \emph{good-3} performance of \emph{Overlap} and the \emph{good-1} performance of plain image retrieval, it becomes apparent that the main part of the clustering gap is due to the object ranking. Hence, there is still room for improved object ranking methods to close this gap.

Fig.~\ref{fig:iconoid_shift_seeds_retrieval_methods} shows diminishing
returns in performance when the number of clusters increases (note the logarithmic x-axis), since the most popular queries are covered
first and the long tail of queries requires exponentially more effort.

The performance gap between \emph{Center} and other methods shows
that the representatives are necessary for ensuring invariance to
different viewing conditions. However, it is desirable to reduce
the set of representatives, since it determines the memory use and
speed of the retrieval index. This is examined more closely in the following section.

\section{Efficient Representations for Retrieval}
\label{sec:retrieval}
Since the discovered landmark representatives are highly redundant,
subsampling them can save memory and computation time.  The goal here
is to reduce the set of representatives in a way that still preserves
as much visual variability as possible in order to ensure good
retrieval performance. The methods we present in the following work by
summarizing groups of similar images, as in \eg
\cite{Avrithis10MM,Gammeter10ECCV}, which can be done efficiently by
exploiting the similarity information that is already available from
the \emph{matching graph} constructed during clustering
\cite{Avrithis10MM,Gammeter10ECCV,Quack08CIVR,Weyand11ICCV,Zheng09CVPR}.
In this graph, every matching pair of images is linked by an edge
whose weight is their matching score. We evaluate four approaches and
compare them against a random baseline. Following
\cite{Avrithis10MM,Gammeter10ECCV,Quack08CIVR} we use the number of
homography inliers as edge weights.

\subsection{Methods}
\label{subsec:compression_methods}
We compare the following five methods for reducing redundancy in the sets of representatives:

\textbf{Complete-Link} (Gammeter \etal \cite{Gammeter10ECCV}) performs hierarchical agglomerative clustering and replaces each complete link component containing at least 3 images by its image with the most neighbors.

\textbf{Kernel Vector Quantization (KVQ)} \cite{Tipping01AISTATS} is a clustering method that selects a minimum number of points such that each point in the dataset is within radius $r$ of at least one selected point. It is used by Avrithis \etal \cite{Avrithis10MM} to reduce the set of features in a cluster after projecting features from all images into a single iconic image, called \emph{Scene Map}. Applying the same idea on the image level, we use it to find a minimal subset of representative images such that each image in a cluster has a given minimum matching score with at least one image in the subset.

\textbf{Dominating Set} chooses a subsample such that each representative is adjacent to at least one image in the original cluster. This subset is found by solving the corresponding set cover problem using the greedy set cover algorithm \cite{Johnson74JCSS}.

\textbf{Fine Iconoids} performs a second, finer Iconoid Shift
\cite{Weyand11ICCV} clustering at bandwidth $\beta=0.7$ that covers
the image collection at a very fine granularity. The representatives
for each (coarse) cluster are then chosen to be all the fine Iconoids
in it. This is similar to the approach of Raguram \etal \cite{Raguram11IJCV} that represents objects by a set of iconic images found by clustering Gist descriptors.

\textbf{Random} is the baseline method that simply draws a random subsample of the representative images.

\subsection{Results}
\begin{figure*}[t]
  \centering 
\hspace{1.5cm}
\includegraphics[width=.4\linewidth]{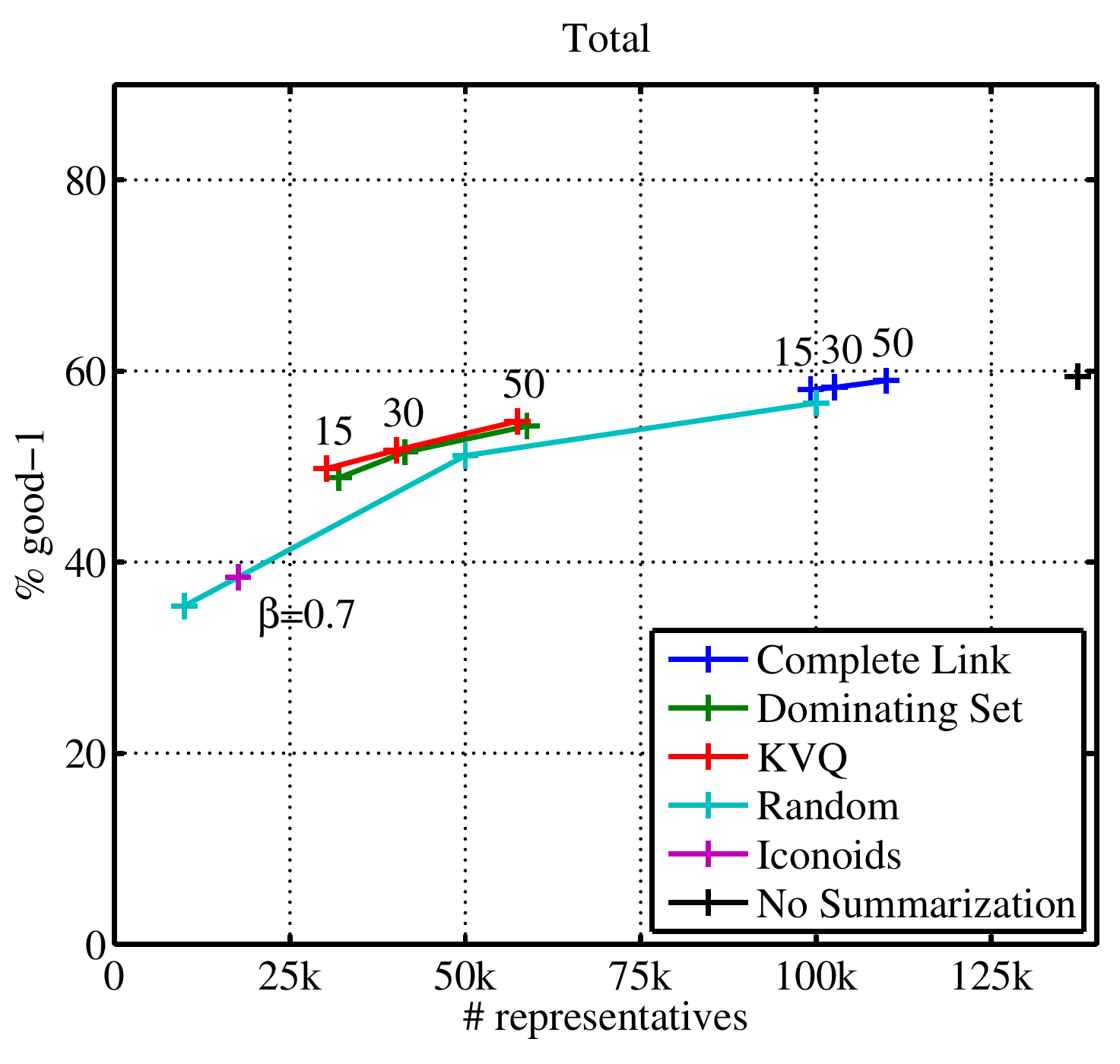}
\hfill
\includegraphics[width=.385\linewidth]{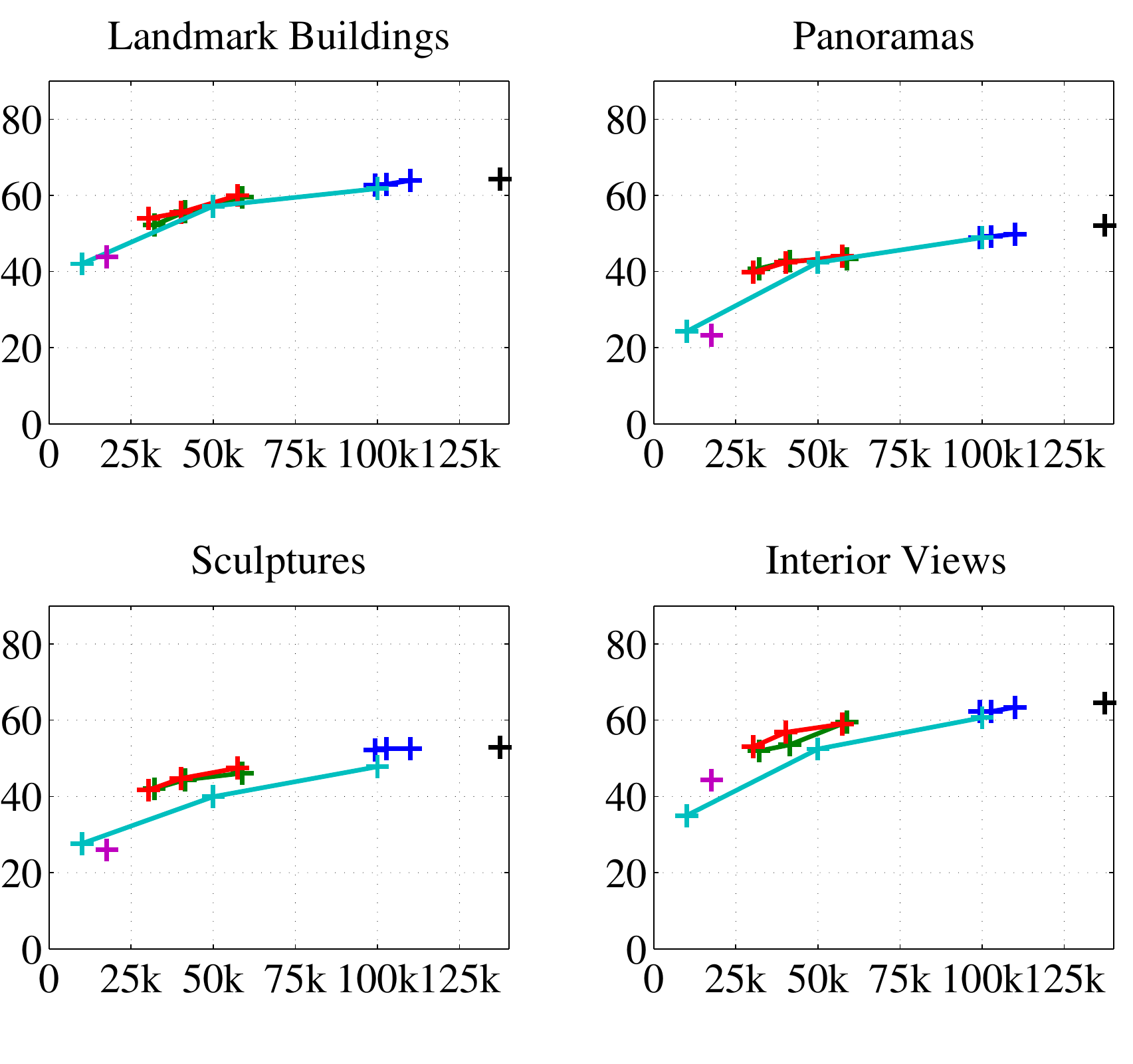}

\includegraphics[width=\linewidth]{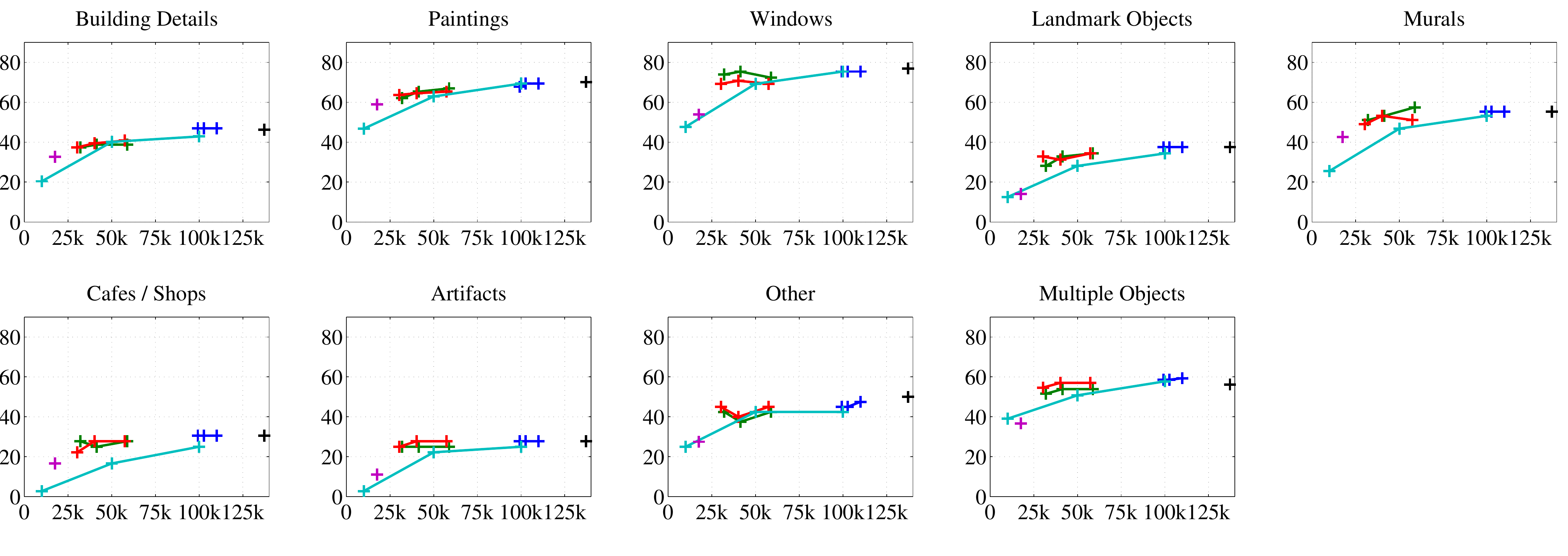}

  \caption{Performance / index size tradeoff of cluster summarization methods using edge weight thresholds of 15, 30 and 50.}
 \label{fig:summarization_methods}
\end{figure*}
We now compare the tradeoffs between the number of representatives and
recognition performance of these methods using the \emph{Voting} method
(Sec.~\ref{subsec:retrieval_methods}) for object scoring
(Fig.~\ref{fig:summarization_methods}). The number of representatives
that the \emph{Complete-Link}, \emph{KVQ} and \emph{Dominating Set}
methods return can be controlled by first deleting edges below a
certain edge weight threshold to make the matching graph sparser and
then running the algorithm. We generated 3 sets of representatives
with each method by applying thresholds of 15, 30 and 50 inliers.

As reported in \cite{Gammeter10ECCV}, \emph{Complete Link} only
slightly reduces the representative set while maintaining high
recognition performance. \emph{Dominating Set} and \emph{KVQ} yield
comparable results since they optimize similar criteria. They
represent a good tradeoff, allowing for a reduction to about 40\%
while still achieving a 60.9\% \emph{good-1} performance (at threshold 50).
An interesting result is that \emph{Fine Iconoids} performs well on 
\emph{Paintings} and \emph{Building Details}, but lower than
\emph{Random} on \emph{Landmark Buildings}. Visual inspection showed that
Iconoids do not cover the more obscure viewing conditions necessary
for robust recognition, since the algorithm is designed to converge to
popular views. It therefore performs higher on categories with a
limited range of possible viewing conditions.

\subsection{Discussion}
In this evaluation, we focused on reduction methods that work on the
\emph{image level}. We have shown that in particular \emph{KVQ} and
\emph{Dominating Set} methods can achieve high compression at only a
small loss in precision.  Some recent approaches also perform this
reduction on the feature level, usually combined with \emph{offline
  query expansion}, \ie, projecting features into matching images,
which is reported to even improve precision over baseline retrieval
\cite{Avrithis10MM,Turcot09LAVD}. An evaluation of these methods would be an
interesting task for future work.

\section{Interfacing Images with Semantics}
\label{sec:semantics}
To find suitable descriptions for the discovered objects, and thus to
enable linking them with information on the web, the usual method is
to perform statistical analysis of the tags and titles that users
provided for the photos in a cluster
\cite{Crandall09WWW,Quack08CIVR,Simon07ICCV,Zheng09CVPR}.
Quack \etal \cite{Quack08CIVR} mine \emph{frequent itemsets} in all tags of a
cluster to generate candidate names. The top-15 candidates are then
used to query Wikipedia, and the retrieved articles are verified by
matching the images occurring in them against the images in the
cluster. In a small informal experiment we found that frequent
itemsets returns many noisy and non-descriptive names like
``vacation'', ``photo'', ``canon'', or ``europe'', which need to be
filtered by a comprehensive stoplist. Furthermore, tags that are
frequently used by the same user like ``summer vacation 2008'' can be ranked higher than correct but less frequent terms.

The method of Simon \etal \cite{Simon07ICCV} is specifically designed
to handle both of these problems. It probabilistically computes a
score $score(c,t)$ for each pair of tag $t$ and cluster $c$. This
score is based on the conditional probability of cluster $c$ given tag
$t$, resulting in tags that mainly occur in cluster $c$. By
marginalizing over the users, tags that are frequent in the
cluster, but used by only few users are ranked low.
We use the method of Simon \etal \cite{Simon07ICCV} in our evaluation
since we found that it yields more reliable tags than the method of
Quack \etal \cite{Quack08CIVR}. We analyze for which objects we can
reliably find semantics and examine the performance gap between object
retrieval and semantic annotation.

\subsection{Data Preparation}
We first need to define a set of tags for each image based on the
metadata provided by the respective photo sharing website. The
\textsc{Paris 500k} dataset consists of images from Flickr and
Panoramio. Since images on Flickr have both tags and a title, we
treated the title as an additional tag. Since images on Panoramio have
no tags, we used the titles as their sole tag. We preprocessed image
tags by applying a very small stoplist containing terms such as
``Paris'' and ``France'' and removing filenames like
``DSC002342.JPG''.

\subsection{Tag Quality Annotation}
\label{subsec:tag_quality_annotation}
In order to analyze tag quality, we manually rated the quality of the
top 3 tags for each object cluster containing 6 or more images.  On
average, annotation of a single object-tag pair took about 30-60
seconds, since often a web search was necessary to verify the
correctness of a tag.  This annotation was performed by six people who
annotated a total of 2,536 objects. The annotators were asked to rate
each image-tag pair as ``good'' if the tag accurately describes what
is visible in the iconic image (\eg, the full name of a building, the
title and painter of a painting) and as ``ok'' if it provides at least
some helpful information, such as the creator of a sculpture, but not
its name, or if the tag is accurate, but contains noise terms like
``me in front of Notre Dame''. This annotation allows us to re-use the
evaluation measures we used for object retrieval in
Sec.~\ref{sec:recognition}.

\begin{figure}[t]
  \centering \includegraphics[width=.49\linewidth]{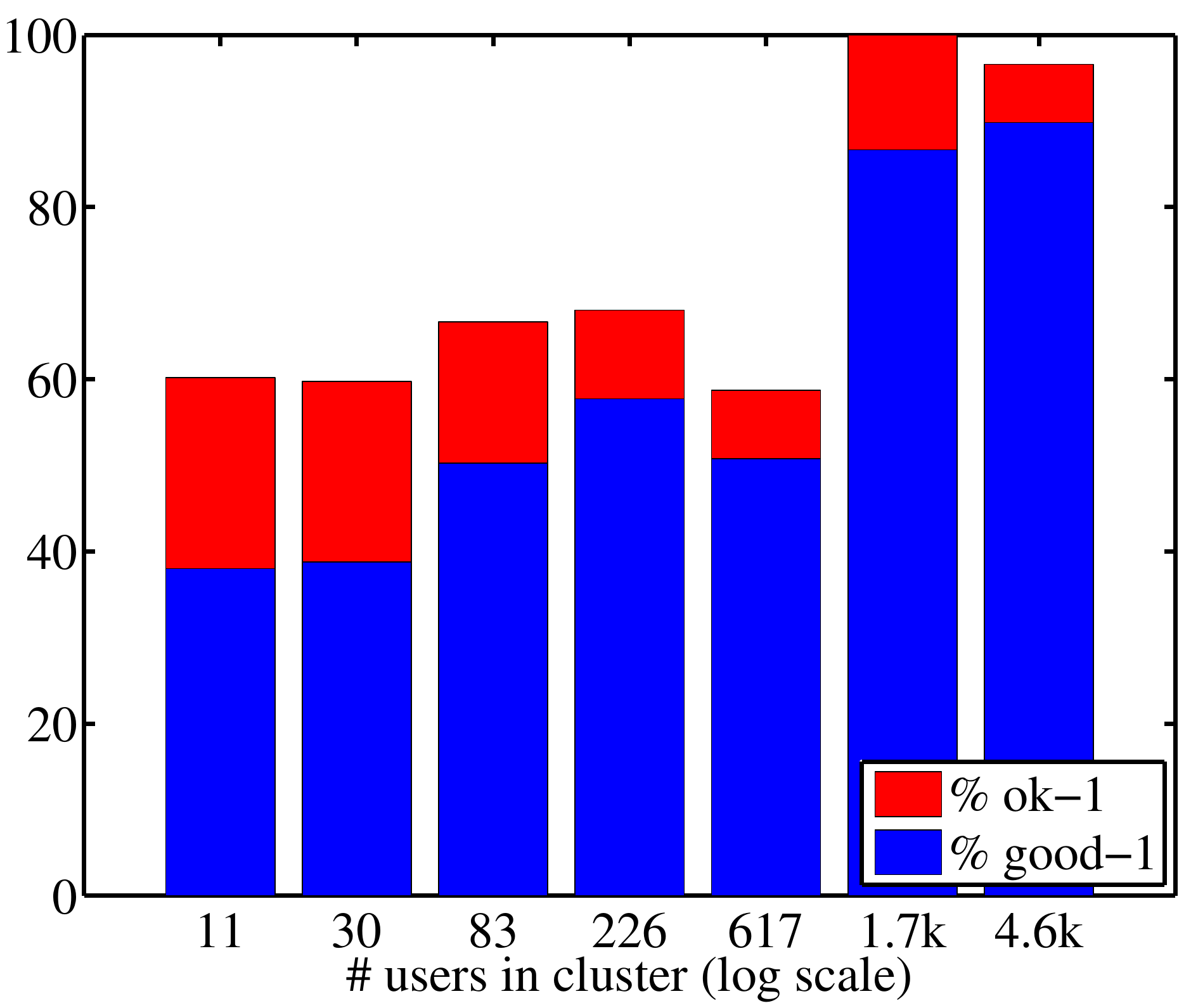}
  \includegraphics[width=.49\linewidth]{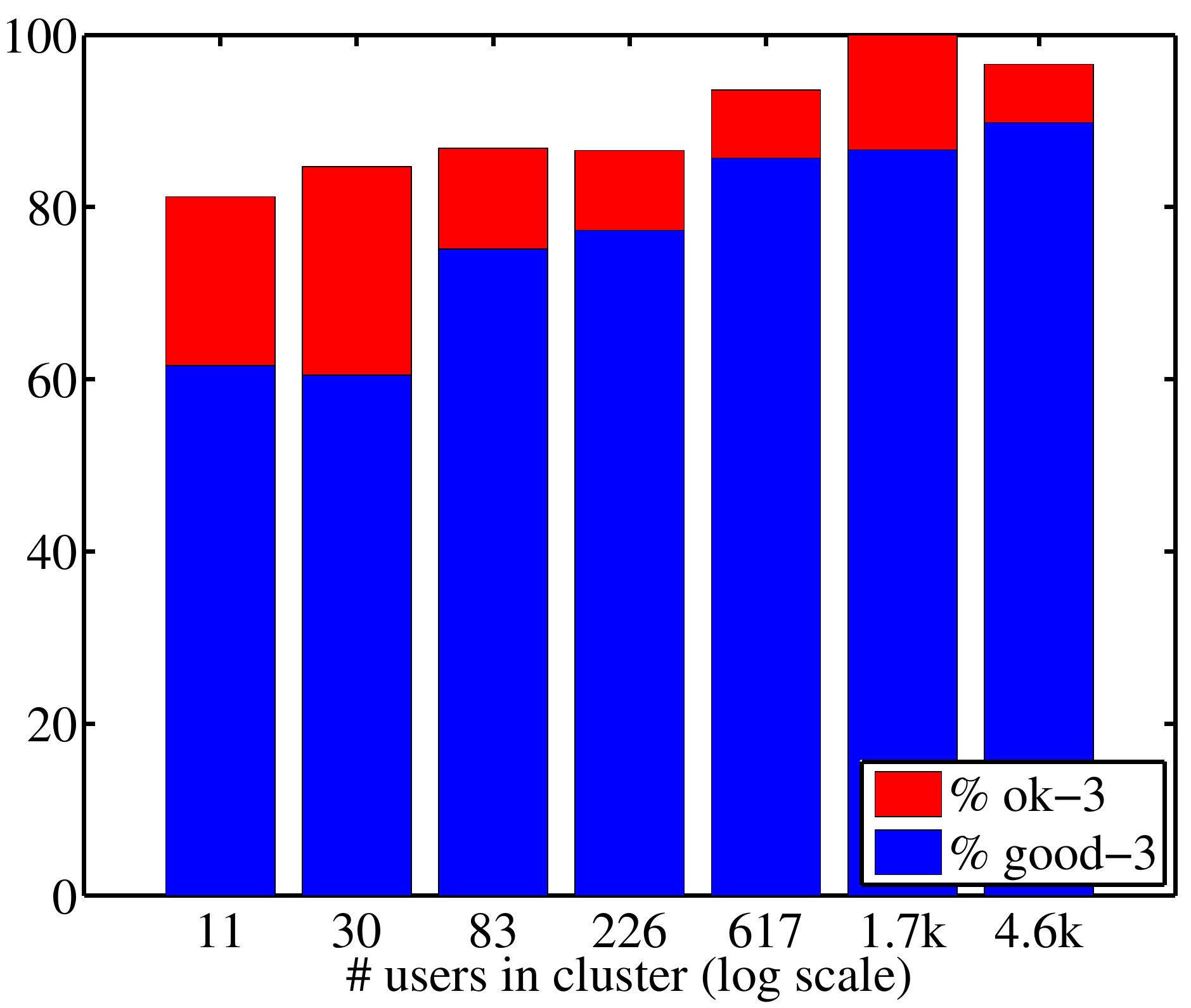}
 \caption{Tag quality as a function of the number of individual users in the cluster. Left: top-1 tag, right: top-3 tags}
  \label{fig:tag_quality_over_cluster_size}
\end{figure}

\begin{figure}[t]
  \centering
    \subfloat[top 1]{
      \includegraphics[width=.6\linewidth]{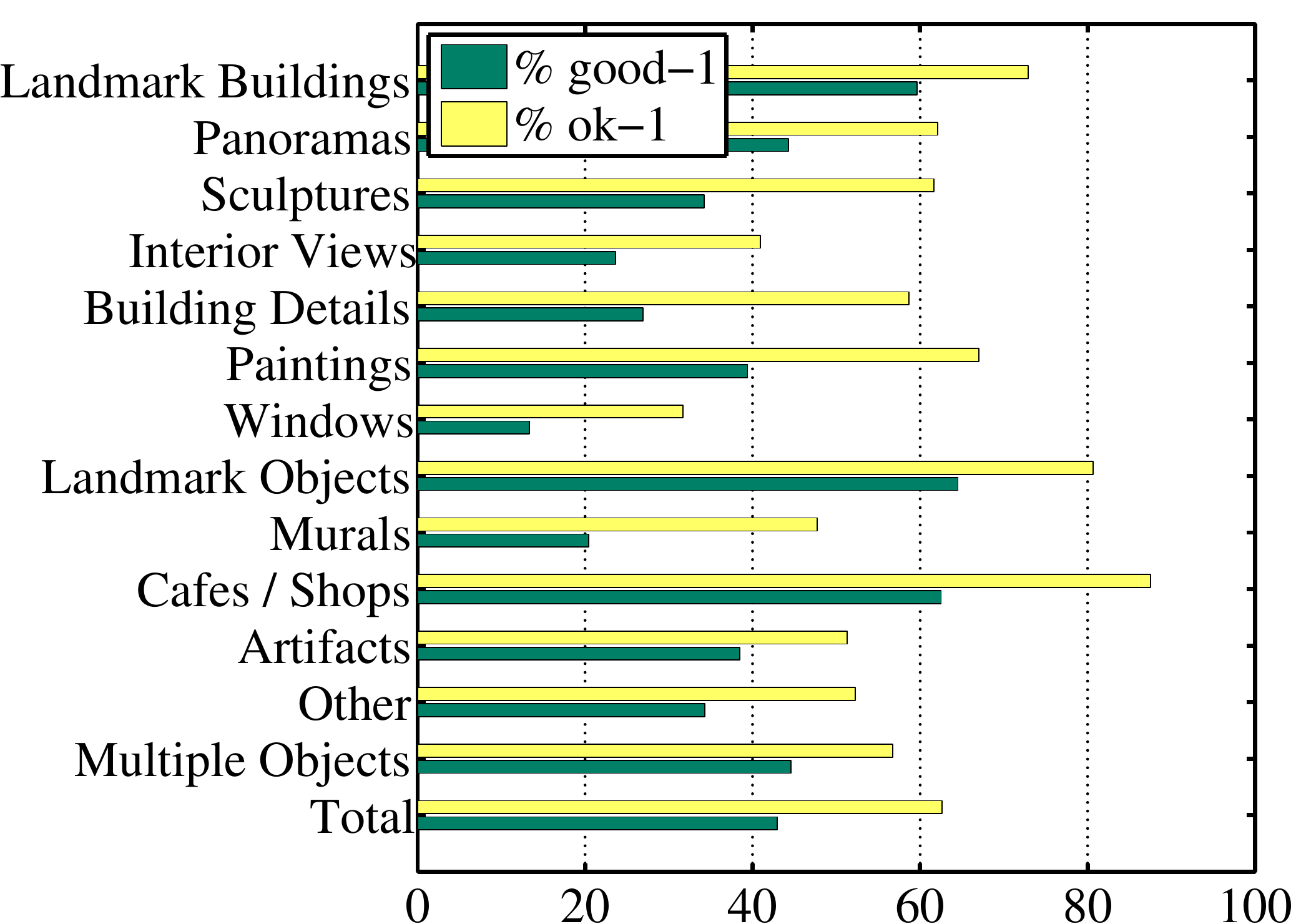}
      \label{fig:tag_quality_iconoid_categories_top_1}
    }
    \subfloat[top 3]{
      \includegraphics[width=.4\linewidth]{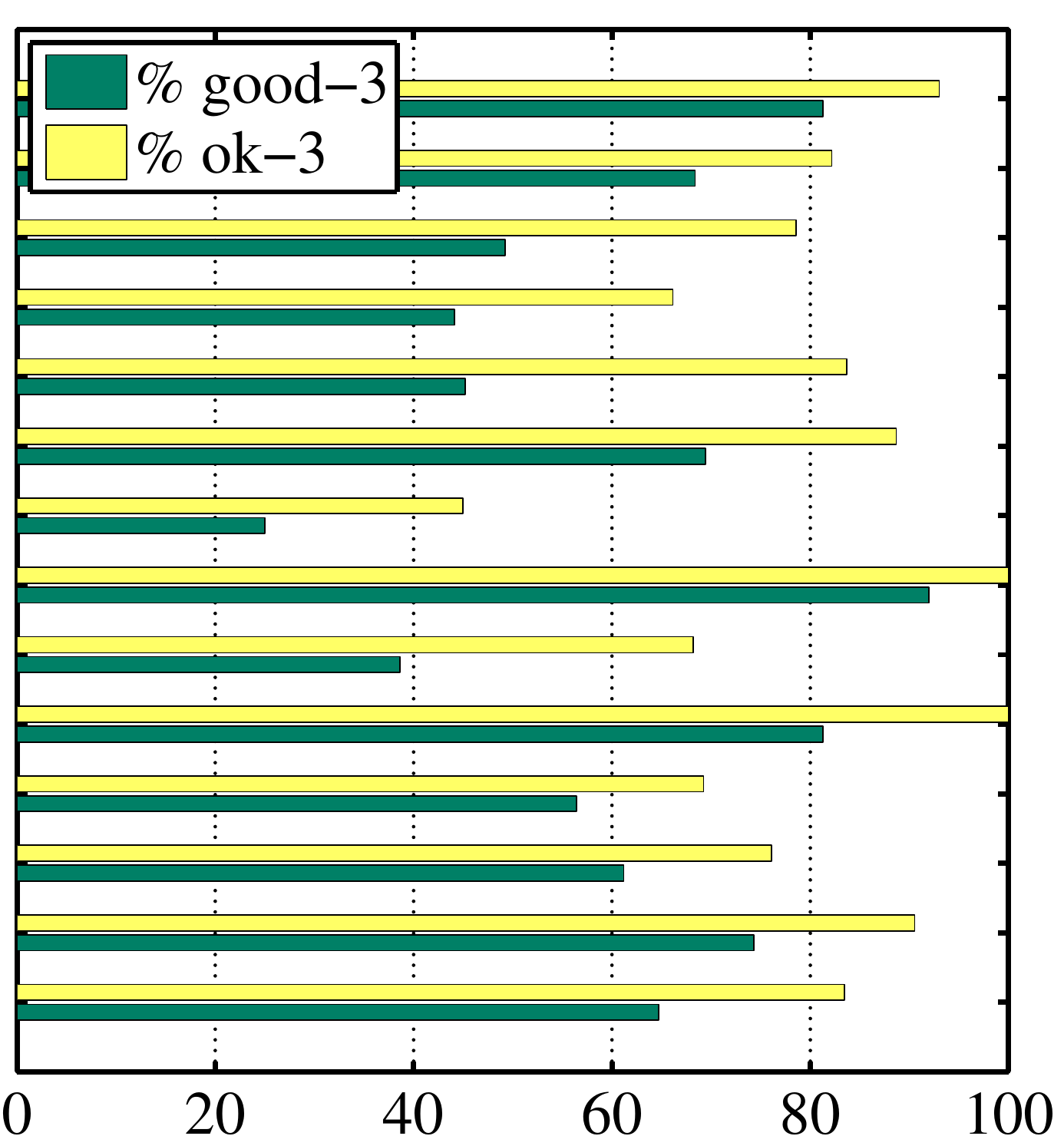}
      \label{fig:tag_quality_iconoid_categories_top_3}
    }
  \caption{Tag quality of different object categories for Iconoid clusters of size 6 and higher.}
  \label{fig:tag_quality_iconoid_categories}
\end{figure}
%

\subsection{Tag Mining}
We first analyze the influence of the number of users contributing
photos to a cluster on the reliability of automatic semantic
annotation.  Simon \etal's method \cite{Simon07ICCV} outputs a ranking
of potential names for each cluster, allowing us to examine the
accuracy of the top-1 and the top-3
tags. Fig.~\ref{fig:tag_quality_over_cluster_size} shows that tag
reliability clearly increases with more users. In particular, over
80\% of clusters with over 1k users have a \emph{good} top-1 tag. With
increasing user count, descriptions also become more precise, since
the fraction of \emph{ok} tags decreases.
Fig.~\ref{fig:tag_quality_iconoid_categories} shows the tag quality
for different categories.  The tags determined for \emph{Landmark
  Buildings}, \emph{Landmark Objects} and \emph{Cafes / Shops} are
most reliable, since their names are typically well-known. \emph{Cafes
  / Shops} are particularly easy to tag since their name is usually
directly visible. \emph{Murals}, \emph{Windows}, \emph{Sculptures} and
\emph{Building Details} are usually lacking proper annotations since
photographers often do not know their names and only label them with
generic tags.  For example, \emph{Building Details} are often tagged
with the name of the entire building. This causes the large difference
in the \emph{good-1} and \emph{ok-1} scores of these categories.

\subsection{Discussion}
\begin{figure}[t]
  \centering
  \includegraphics[width=\linewidth]{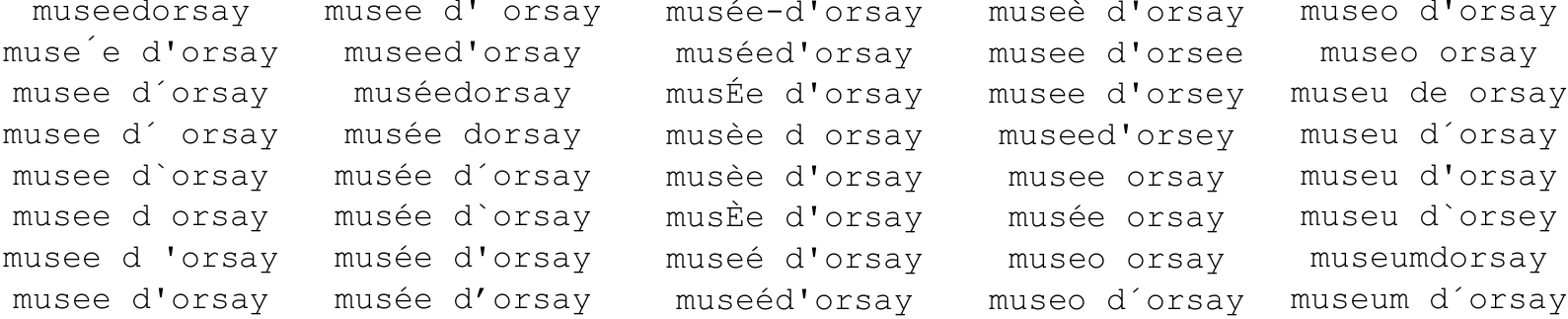}
  \caption{Different spellings of ``Musée d'Orsay'' encountered in the dataset.}
  \label{fig:musee_d_orsay_spellings}
\end{figure}
While for most large clusters suitable semantics can be found, for
small and medium sized clusters the difference in \emph{ok-1} and \emph{good-1}
scores (Fig.~\ref{fig:tag_quality_over_cluster_size}) suggests that
better tag ranking could greatly help the recognition of less popular
objects. Significant improvements can likely be made by increasing
robustness \wrt different languages, spelling errors and tag
noise. For example, Fig. \ref{fig:musee_d_orsay_spellings} shows a
selection of different spellings of ``Musée d'Orsay'' from our dataset
that would all be treated as separate tags by current methods. Mining
Wikipedia \cite{Quack08CIVR} or tourist guide websites
\cite{Zheng09CVPR}, or performing specialized per-category metadata
mining as done by Arandjelović \etal \cite{Arandjelovic12ICMR} for
sculptures might also help in naming the less popular objects.

\section{End-to-End Analysis}
\label{sec:end_to_end}
\begin{figure}[t]
  \centering
  \includegraphics[width=\linewidth]{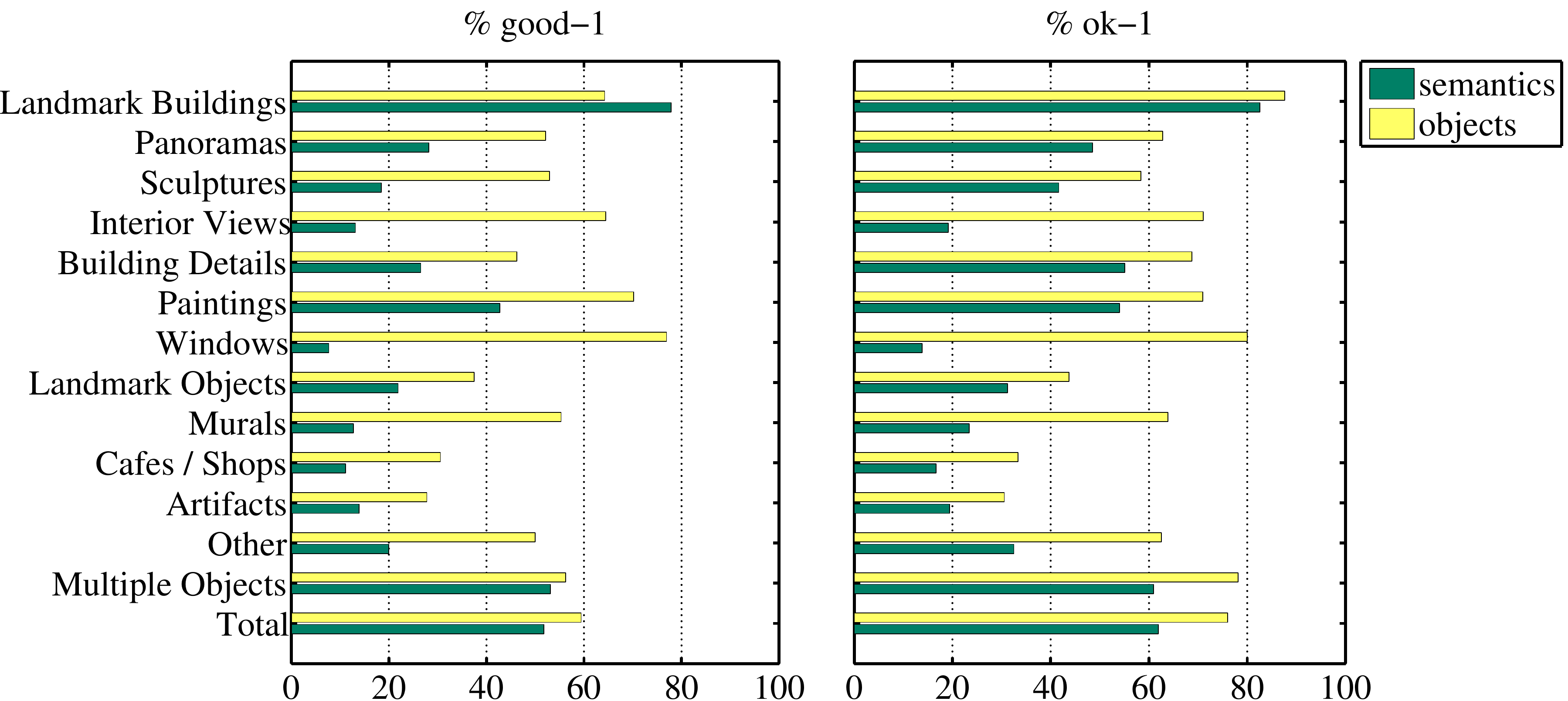}
  \caption{Quality of good (left) and ok (right) semantics assigned to queries, compared to the quality of the objects retrieved.}
  \label{fig:query_tag_assignment_quality_for_different_query_types}
\end{figure}
In Sec.~\ref{sec:recognition}, we analyzed different methods for assigning \emph{objects} to queries and measured accuracy on a visual level. In this section, we perform this analysis on a  \emph{semantic} level, based on the labels assigned in the previous section.

\subsection{Setup}
To evaluate the system from end to end, we cluster the \textsc{Paris500k}
dataset with \emph{Iconoid Shift} using 100k seeds and mine semantics
for the clusters using the method of Simon \etal
\cite{Simon07ICCV}. We use the \emph{Voting} object scoring method
(Sec.~\ref{sec:recognition}) to rank objects \wrt a query. Subsampling
(Sec.~\ref{sec:retrieval}) is not used.  We then rate the relevance of
the top-scoring tag of the top-scoring object for each query as either
\emph{good}, \emph{ok} or \emph{bad}
(Sec.~\ref{subsec:tag_quality_annotation}).

\subsection{Results}
Fig.~\ref{fig:query_tag_assignment_quality_for_different_query_types}
shows the results of semantic annotation and compares it to the plain object
recognition performance (yellow bars in Fig.~\ref{fig:retrieval_methods_landmark_types}).
\emph{Landmark Buildings} have the highest performance, because their
large cluster size enables robust recognition and semantic
assignment. The reason semantic assignment has even higher performance
than plain object recognition is that semantic assignment sometimes
corrects errors of object recognition: It often happens that the query
image shows the whole building, but the matching object is a detail of
the building, \eg a door of Notre Dame or a leg of the Eiffel
Tower. This occurs, \eg, because the detail is most prominent in the
query due to perspective, or because large parts of the building are
occluded, but the detail is still visible. However, photos of building
details are often labeled with the name of the whole building, since
photographers do not know, \eg, the name of a particular door of Notre
Dame. Therefore, the query is correctly assigned the name of the whole
building even though the recognized object was a detail.

For categories with smaller clusters, there are different bottlenecks: \emph{Artifacts} and
\emph{Landmark Objects} are hard to recognize, because they are compact
objects and form only small clusters. However, they typically have
high quality tags (Fig.~\ref{fig:tag_quality_iconoid_categories}).
In other cases, a relevant object can reliably be retrieved, but low
tag quality prevents successful semantic annotation.  This problem is
most prominent for \emph{Windows} and \emph{Murals}
(Fig.~\ref{fig:query_tag_assignment_quality_for_different_query_types}). They
are flat and photographed under a limited range of conditions, making
them easy to recognize, but they suffer from low quality semantics,
since information on them is not easily available.
In total, 51.8\% of the queries could be assigned
good semantics, while for 59.4\% a good object was retrieved;
61.9\% of queries were assigned ok tags, while for 76.0\% an ok object
was retrieved. In the following, we summarize the causes of this gap and discuss possible solutions.

\section{Discussion and Conclusion}
\label{sec:discussion}
We now sum up the findings of our evaluation by answering the questions posed in the introduction and discuss the areas where improvements can still be made.

\subsection*{How many and what kinds of objects are present in Internet photo collections and what is the difficulty
of discovering objects of different object categories?}
The question how many objects there are cannot be answered based directly on the number of clusters, because there can be multiple clusters of the same object showing different views, and clusters of non-objects, \eg party photos, pictures of animals and food, or photo bursts. We performed an annotation experiment (Sec.~\ref{sec:clusters_per_cat}) and labeled the 3,088 clusters containing five or more images discovered in the \textsc{Paris500k} dataset. 2,585 (83.7\%) clusters were labeled as objects and 503 (16.3\%) were labeled as non-objects (Note that this ratio will shift more strongly towards non-objects when also considering clusters containing less than 5 images.) Approaches for detecting and removing such non-object clusters could be a direction for future work, since they unnecessarily increase the size of the retrieval index and increase the chance of recognition errors. The distribution of object categories (Fig.~\ref{fig:iconoid_colormatrices}) shows that, not surprisingly, \emph{Landmark Buildings} from the largest category, followed by \emph{Sculptures}, \emph{Panoramas} and \emph{Paintings}. Seed-based object discovery algorithms \cite{Weyand11ICCV,Chum10PAMI,Chum09CVPR} find the most photographed objects first, because when drawing a random image from an Internet photo collection, the likelihood of drawing an often photographed object is higher. Therefore, much more effort is required to also discover objects in the long tail of the size distribution. This could be addressed in future work, \eg by seeding methods that avoid the bias to large clusters, or methods that explicitly mine for small objects \cite{Chum09CVPR,Letessier12ACMMM}.

\subsection*{How to decide which landmark was recognized given a list of retrieved images?}
We analyzed five methods for this task (Sec.~\ref{sec:recognition}). Our experiments clearly showed that having a set of representative images for each cluster is necessary to recognize it under difficult conditions such as extreme view differences, occlusion, lighting changes, blur, etc.\ (first two rows of Fig.~\ref{fig:query_iconoid_tags_good}). This can be viewed as a form of offline query expansion. However, like query expansion, this method is also prone to \emph{drift} (Fig.~\ref{fig:query_iconoid_tags_bad}, top row), which can cause confusion between nearby objects. While the often used \emph{Best Match} method \cite{Quack08CIVR,Zheng09CVPR,Gammeter10ECCV,Gammeter09ICCV} avoids drift better than some other methods, we found that by using a method that explicitly maximizes the \emph{Overlap} between the query and the object's iconic image, even higher precision can be achieved. However, the ranking gap (\emph{good-1} vs. \emph{good-3} performance) remains relatively large, suggesting that there is potential for more accurate object ranking methods.

\subsection*{How to efficiently represent the discovered objects in memory for recognition?}
We analyzed four \emph{image level} techniques for eliminating redundancy in the database (Sec.~\ref{sec:retrieval}). Our analysis revealed that it is important to keep representatives showing obscure views and extreme lighting conditions. Therefore, representing the object by a set of popular views (as done in our \emph{Fine Iconoids} method or in \cite{Raguram11IJCV}) did not perform better than random subsampling. We proposed two methods that achieve an acceptable tradeoff between database size and recognition performance, but observed that there is still potential for better methods.
A comparative analysis of methods that eliminate redundancy at the \emph{feature level} \cite{Avrithis10MM,Gammeter09ICCV,Turcot09LAVD} would also be an interesting future direction.

\subsection*{Are the user-provided tags reliable enough for determining accurate object names?}
\begin{figure}[t]
  \centering
  \includegraphics[width=\linewidth]{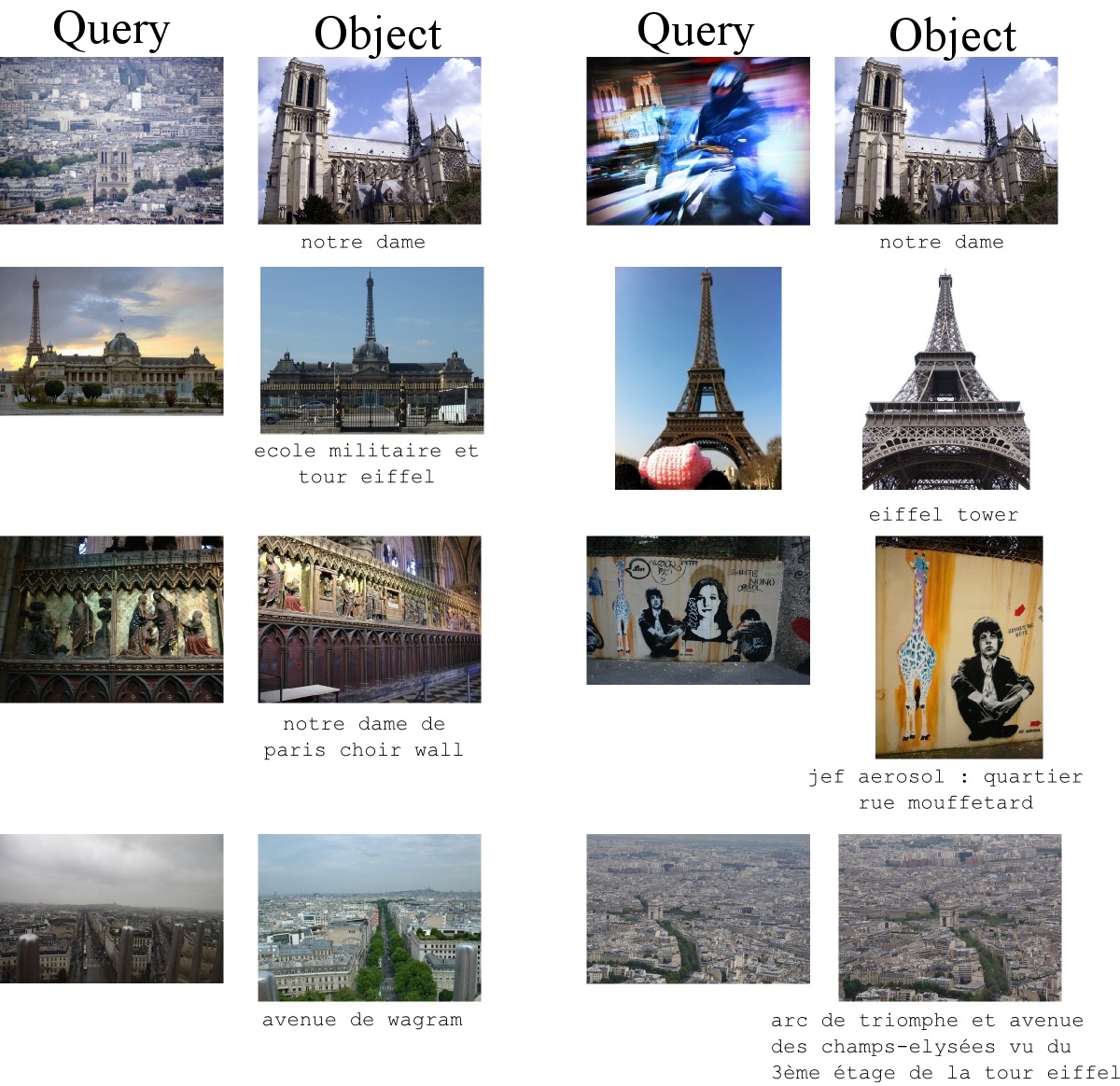}
  \caption{Examples of successful recognition and annotation.}
  \label{fig:query_iconoid_tags_good}
\end{figure}
The largest room for improvement of the overall performance of
landmark recognition is in semantic annotation. We determined two
reasons for the performance loss at this step
(Sec.~\ref{sec:semantics}): (1) User-provided tags come in different
languages, have spelling errors and contain noise terms (Fig.~\ref{fig:musee_d_orsay_spellings}). Methods
robust to these factors could provide more reliable semantics even for
small clusters. (2) Often, insufficient information is available to
photographers, causing non-descriptive tags
(Fig.~\ref{fig:query_iconoid_tags_bad}, bottom right). This might be
addressed by crawling relevant encyclopedia \cite{Quack08CIVR} or tourist guide articles \cite{Zheng09CVPR} or using image search engines \cite{Arandjelovic12ICMR}.
However, sometimes, the presence of accurate tags in small
clusters can also lead to surprisingly accurate results
(Fig.~\ref{fig:query_iconoid_tags_good}, rows 3 and 4).

\subsection*{What are the factors effectively limiting the recognition of different landmark types?}
\begin{figure}[t]
  \centering
  \includegraphics[width=\linewidth]{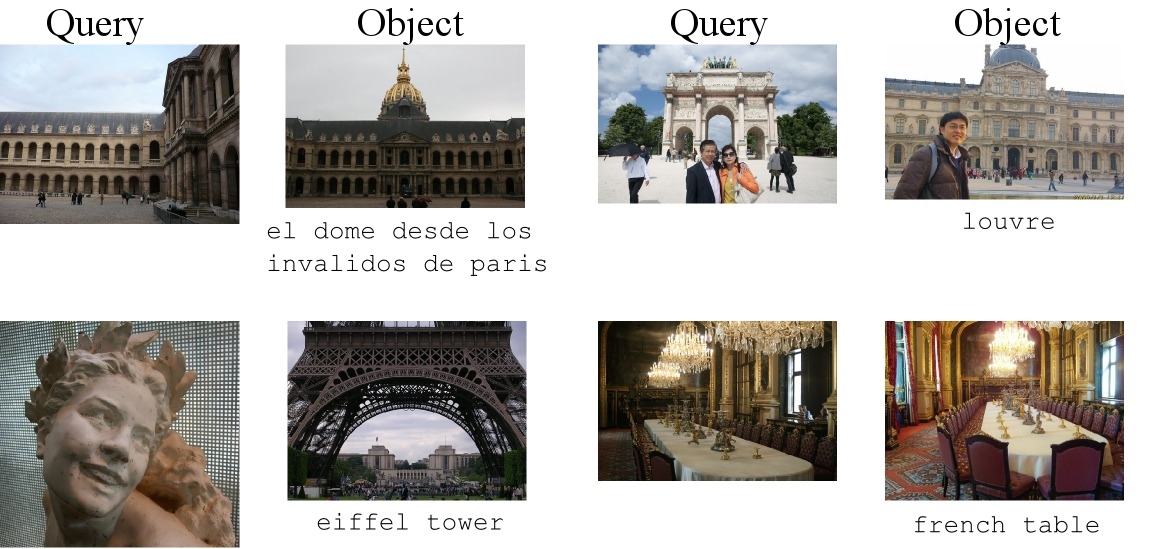}
  \caption{Examples of failure modes. Top row: Drift due to dominance
    of nearby objects. Bottom left: Failed retrieval caused by repeating
    patterns. Bottom right: Too generic description.}
  \label{fig:query_iconoid_tags_bad}
\end{figure}

Our end-to-end analysis (Sec.~\ref{sec:end_to_end}) showed that the factors that limit recognition performance are quite varied and strongly depend on the object category. Some objects like \emph{Windows} or \emph{Murals} are easy to recognize \emph{visually}, but often lack accurate tags, which prevents semantic assignment. This could be addressed using the methods mentioned above. Other objects such as \emph{Artifacts} or \emph{Landmark Objects} do have accurate tags, but are harder to recognize due to their spatial structure and small cluster size. Their recognition could be improved by mining more photos of them from the web \cite{Gammeter10ECCV}.
Finally, improvements to image retrieval and matching, \eg improved feature representation \cite{Arandjelovic12CVPR}, improved feature quantization \cite{Jegou08ECCV,Jegou11PAMI}, or
ranking methods robust to problems like repeated patterns \cite{Jegou09CVPR} (Fig.~\ref{fig:query_iconoid_tags_bad}, bottom left), will directly benefit both landmark clustering and recognition.

\subsection{Limitations}
While our evaluation has brought to light several opportunities for
progress, its scope could still be broadened in future work.
Due to our choice of dataset our taxonomy of queries is certainly
biased towards the landmarks of Paris. A larger dataset from several
cities would increase the generality of the evaluation.
Our query set was collected from Internet photo collections and is
therefore representative for the task of photo auto-annotation. While
this bias only affects the score average and not the per-category
scores, a second query set for the task of mobile visual search would make it possible to identify problems specific for that task.
The set of methods we analyzed was carefully chosen, but an analysis
of other approaches (\eg, for clustering or semantic
annotation) may bring further insights into how the component choices 
affect overall performance.

\subsection{Conclusion}
In this work, we have evaluated the automatic construction of visual
landmark recognition engines from Internet image collections. We used
a large-scale dataset of 500k photos from Paris, collected a set of 3k
typical query images, and created a ground truth for evaluating
large-scale photo auto-annotation that we made publicly available. For
each component of the pipeline, we evaluated how different methods and
parameters affect overall performance as well as the performance for
individual query categories. We proposed several novel methods for
various sub-tasks, some of which outperform literature approaches. In
our analysis, we have identified areas where such a system performs
well, as well as areas where improvement is still possible.

\section*{Acknowledgments}
This project has been funded, in parts, by a Google Faculty Research Award and by the cluster of excellence UMIC (DFG EXC 89).

\bibliographystyle{model1-num-names}
\bibliography{abbrev,weyand_landmark_recognition}

\end{document}